\documentclass{article} 
\usepackage{iclr2026_conference,times}

\usepackage{amsmath,amsfonts,bm}

\def\eqref#1{equation~\ref{#1}}

\def\1{\bm{1}}

\DeclareMathAlphabet{\mathsfit}{\encodingdefault}{\sfdefault}{m}{sl}
\SetMathAlphabet{\mathsfit}{bold}{\encodingdefault}{\sfdefault}{bx}{n}

\usepackage{hyperref}
\usepackage{url}
\usepackage{graphicx}
\usepackage{svg}
\usepackage[table]{xcolor}
\usepackage{colortbl}
\usepackage{multicol}
\usepackage{multirow}
\usepackage{algorithm}
\usepackage{algpseudocode}
\usepackage{booktabs}
\usepackage{fontawesome5}
\usepackage{subcaption} 

\definecolor{gray}{rgb}{0.8, 0.8, 0.8}
\usepackage{float} 
\definecolor{lightred}{RGB}{250, 200, 200}

\usepackage{tikz}

\usepackage{wrapfig}
\usepackage[most]{tcolorbox}
\usepackage{xcolor}

\title{ssToken: Self-modulated and Semantic-aware Token Selection for LLM Fine-tuning}

\author{Xiaohan Qin$^{1,2}$, Xiaoxing Wang$^{1}$, Ning Liao$^{1}$, Cancheng Zhang$^{1}$, Xiangdong Zhang$^{1}$, \\
\textbf{Mingquan Feng}$^{1}$, \textbf{Jingzhi Wang}$^{1}$, \textbf{Junchi Yan}$^{1,2,\dag}$\\
$^{1}$Shanghai Jiao Tong University, $^{2}$Shanghai Innovation Institute\\
\url{https://github.com/jianke0604/ssToken}
}

\definecolor{metacolor}{HTML}{0064E0}
\hypersetup{
    colorlinks=true,
    citecolor=metacolor,
}

\let\cite\citep
\iclrfinalcopy 
\begin{document}

\maketitle

\begin{abstract}
Data quality plays a critical role in enhancing supervised fine-tuning (SFT) for large language models (LLMs), and token-level data selection has emerged as a promising direction for its fine-grained nature. Despite their strong empirical performance, existing token-level selection methods share two key limitations: (1) requiring training or accessing an additional reference model, and (2) relying solely on loss information for token selection, which cannot well preserve semantically important tokens that are not favored by loss-based metrics. To address these challenges, we propose \textbf{ssToken},  a \textbf{S}elf-modulated and \textbf{S}emantic-aware \textbf{Token} Selection approach. ssToken leverages readily accessible history models to compute the per-token loss difference with the current model, which serves as a self-modulated signal that enables the model to adaptively select tokens along its optimization trajectory, rather than relying on excess loss from an offline-trained reference model as in prior works. We further introduce a semantic-aware, attention-based token importance estimation metric, orthogonal to loss-based selection and providing complementary semantic information for more effective filtering. Extensive experiments across different model families and scales demonstrate that both self-modulated selection and semantic-aware selection alone outperform full-data fine-tuning, while their integration—ssToken—achieves synergistic gains and further surpasses prior token-level selection methods, delivering performance improvements while maintaining training efficiency. 
\end{abstract}

\section{Introduction}
\label{sec:intro}
Supervised fine-tuning (SFT) has emerged as a crucial stage in the modern training pipeline of large language models (LLMs), enhancing their instruction-following capability and practical utility.
While early efforts primarily focused on scaling up instruction-tuning corpora, sometimes with millions of samples~\cite{wang2022self,longpre2023flan,chung2024scaling}, recent findings have consistently demonstrated that \textit{data quality outweighs data quantity} in SFT~\cite{chen2023alpagasus,zhou2024lima,fu2025t}, as small but carefully curated datasets often yield stronger downstream performance than much larger but noisier collections. 
Therefore, identifying and prioritizing subsets of data that contribute most to model improvement from massive candidate data pools has become a key challenge in LLM supervised fine-tuning.

Recent studies~\cite{rho,fu2025t,simoulin2025memory} have shown that even after rigorous sample-level filtering, high-quality datasets still contain substantial token-level noise that degrades training outcomes. 
Non-task-related patterns or phrases are often redundant or uninformative, and continuing to fine-tune on them offers limited benefit while potentially undermining downstream task performance.
This has motivated growing interest in token-level data selection, which aims to refine training datasets at a finer granularity by filtering individual tokens rather than entire samples. Importantly, token-level approaches are orthogonal to domain-level and sample-level strategies, thereby broadening their applicability across diverse data selection scenarios.

Despite its advantages, we observe that existing token-level selection methods~\citep{rho,tokencleaning} share two key limitations: (1) requiring training or accessing an additional reference model, and (2) relying solely on loss information for token selection, which cannot well preserve semantically important tokens that are not favored by loss-based metrics.
Specifically, recent approaches typically prepare a task-related high-quality dataset and fine-tune either the base model or a smaller proxy model on it to serve as the reference, or alternatively adopt a more powerful model with the same tokenizer guided by prior knowledge. Excess loss is then computed against this reference model to identify task-related tokens that are important for improving downstream performance and instruction-following capability, as well as learnable tokens whose loss steadily decreases.
However, directly adopting a stronger model is not always practical, and training an extra reference model incurs additional time and resource costs. 
Moreover, \cite{tokencleaning} has shown that the performance of the reference model itself has a substantial impact on the effectiveness of token selection.
On the other hand, while token-level losses reflect the model’s prediction uncertainty, they do not necessarily capture the semantic importance of tokens within their specific context. Frequent yet semantically uninformative tokens may yield similar excess loss values as task-critical ones, making loss-only selection prone to discarding informative content.

\begin{figure}[t!]
    \centering
    \includegraphics[width=\linewidth]{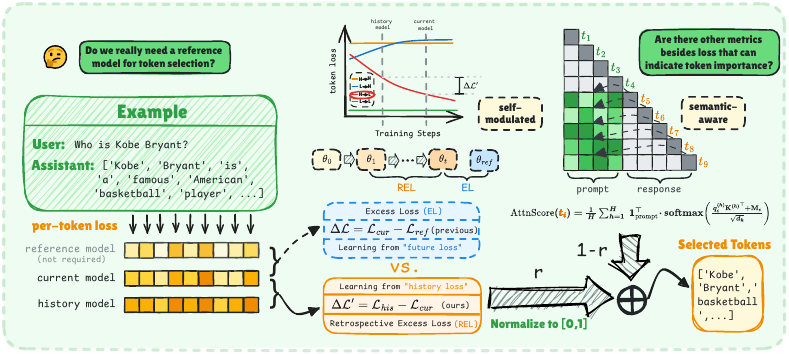}
    \caption{The overall framework of ssToken, which consists of two key components: (i) the self-modulated token selection strategy and (ii) the semantic-aware token importance estimation. We also provide a comparison with prior reference–model–based methods.}
    \label{fig:workflow}
    \vspace{-15pt}
\end{figure}

In this paper, we propose \textbf{ssToken} (\textbf{S}elf-modulated and \textbf{S}emantic-aware \textbf{Token} Selection), a new token-level data selection method designed to address the two limitations discussed above. 
The core role of a reference model is to identify task-related and learnable tokens while discarding noisy or already well-learned ones. Instead of preparing an additional curated dataset and training a separate reference model, we view the current model itself as a natural teacher: as training progresses, its improvements over its own history provide reliable signals for token selection. As shown in Fig.~\ref{fig:workflow}, relative to the history model (e.g., the base model before SFT), the current model has already learned certain task-related patterns and achieved stronger performance, which can be viewed as a reference.
This perspective aligns with the principle of self-modulated learning, where the model adaptively regulates tokens that remain informative and learnable along its trajectory. Intuitively, if the current model achieves a notable loss reduction on a token compared with the history model, that token is unlikely to be noisy or already well-learned, but rather represents learnable and informative content. Furthermore, as the current model improves, the history model can be updated adaptively (e.g., via EMA), providing more stable guidance than relying on a fixed reference model. 

On the other hand, for a well-pretrained model, the attention matrix over input samples inherently encodes rich semantic information, which we regard as a complementary signal beyond loss for assessing token importance. To leverage this property in the fine-tuning setting, we design an attention-based semantic-aware selection metric that identifies tokens more semantically salient in context and more beneficial for enhancing instruction-following capability. Extensive experiments show that both self-modulated selection and semantic-aware selection alone outperform full-data fine-tuning, while their integration—ssToken—achieves synergistic gains and further surpasses previous token-level selection methods. 
Overall, our contributions can be listed as follows:

\textbf{1) Self-modulated token selection paradigm.} We replace costly or even unavailable reference models with a self-modulated strategy that leverages the current model’s improvements over its history, thereby avoiding additional time and resource costs. Unlike a fixed reference model, the history model can be updated throughout training, providing more stable and long-term guidance.

\textbf{2) Semantic-aware token importance estimation metric.} We introduce an attention-based metric that leverages the rich semantic information embedded in attention matrices, offering a complementary signal to loss-based selection. The two signals are orthogonal, and their integration leads to mutually reinforcing effects. Additionally, we design a lightweight implementation that is compatible with efficient attention mechanisms such as FlashAttention, thus introducing negligible computational overhead.

\textbf{3) Strong empirical performance across various model families and scales.} We conduct extensive experiments on models ranging from 3B to 14B parameters across multiple benchmarks. Results consistently show that our proposed ssToken not only surpasses full-data fine-tuning baselines by up to \textbf{4.3\%} but also outperforms previous token-level selection methods by up to \textbf{2.8\%}, while maintaining training efficiency.

\section{Related Work}

\textbf{LLM Data Curation.} The core objective of data curation for LLMs is to maximize training performance and efficiency by enhancing the quality and scale of training data. This encompasses a wide range of strategies, including acquiring data through web crawling~\cite{penedo2024fineweb,yu2025craw4llm} or synthetic generation~\cite{wang2022self,li2023synthetic,long2024llms,han2025attributes,wang2025reverse}, filtering~\cite{gao2020pile,penedo2023refinedweb} and deduplication~\cite{lee2021deduplicating,kandpal2022deduplicating}, rebalancing across domains~\cite{xie2023doremi,liu2024regmix}, selecting subsets from a given data pool~\cite{liu2025quadmix,wang2025data,fu2025t}, and ranking data to enable curriculum learning~\cite{lee2023instruction,chen2023skill}. In this work, we focus on data selection for LLM supervised fine-tuning.

\textbf{LLM Data Selection.} Data selection is a critical component of the LLM data curation pipeline, aiming to further refine filtered and deduplicated high-quality datasets to improve training performance and efficiency~\cite{raffel2020exploring,tan2023data,bukharin2023data,yu2024diversify}. Sample-level selection has been extensively explored by recent studies~\cite{pang2024improving,li2024superfiltering,fu2025t}, and some approaches~\cite{zhang2024speculative,wang2025data} can achieve nearly lossless performance using less than 10\% of the original data for fine-tuning. Recently, \textsc{RHO-1}~\cite{rho} highlights that even carefully curated datasets at the sample level still contain token-level noise and uninformative content. They are the first to explore token-level data selection for LLM training and report substantial gains over full-data baselines. TokenCleaning~\cite{tokencleaning} further optimizes token-level selection in the SFT setting, introducing fixed-model cleaning and self-evolving cleaning strategies. However, these methods all share the two limitations discussed in the introduction. In contrast, our proposed ssToken avoids the need to train or access an extra reference model and leverages attention-based signals to complement loss information with additional semantic cues.

\section{Methods}

\subsection{Preliminaries: Reference-model-based Token Selection}
\label{sec:preliminaries}
Given a data pool $\mathcal{D}$ with $N$ samples, let $\boldsymbol{x}$ = $\{x_1, x_2, \ldots, x_L\} \in \mathcal{D}$ denote one sample with sequence length $L$. Training a large language model  under the next-token prediction paradigm can be formulated as minimizing the negative log-likelihood of observed tokens in the dataset. The loss on a single sample $\boldsymbol{x}$ is defined as
\begin{equation}
\label{eq:slm}
\mathcal{L}_\theta(\boldsymbol{x}) = - \frac{1}{|\mathcal{I}|} \sum_{i \in \mathcal{I}} \log \mathbb{P}_\theta(x_i \mid x_{<i}),
\end{equation}
where $\theta$ denotes the model parameters, $x_i$ is the $i$-th token, and $x_{<i} = \{x_1, \ldots, x_{i-1}\}$ denotes the preceding context. The index set $\mathcal{I} \subseteq \{1, \ldots, L\}$ specifies which tokens are supervised. For pre-training, $\mathcal{I}$ typically includes the entire sequence; for SFT, $\mathcal{I}$ only covers the response tokens, while prompt tokens serve as conditioning context but are not directly supervised.

RHO-1~\cite{rho} observes that when continuing training on a well-pretrained LLM, token losses can be categorized into four types based on their trajectories: persistent high loss (H$\rightarrow$H), increasing loss (L$\rightarrow$H), decreasing loss (H$\rightarrow$L), and consistently low loss (L$\rightarrow$L). 
It prioritizes task-related tokens that the model can stably learn (H$\rightarrow$L), rather than those already mastered (L$\rightarrow$L) or those that remain persistently difficult or noisy (H$\rightarrow$H and L$\rightarrow$H). To this end, a reference model is obtained by further training the base model on a high-quality task-related dataset. 
Token-level \textbf{Excess Loss (EL)} is then computed with respect to this reference model to score and filter tokens. Specifically, the excess loss for token $x_i$ is defined as:
\begin{equation}
\text{EL}(x_i) \;=\; \mathcal{L}_\theta(x_i) - \mathcal{L}_{\theta_\text{ref}}(x_i) 
= \log \frac{\mathbb{P}_{\theta_\text{ref}}(x_i \mid x_{<i})}{\mathbb{P}_\theta(x_i \mid x_{<i})}.
\end{equation}
Here, $\text{EL}(x_i)$ measures the extent to which the prediction probability of token $x_i$ is expected to improve in future training. A larger value indicates that the model is more likely to achieve greater predictive gains on this token. They then rank tokens in the data pool based on EL and their designed scoring strategies, followed by token selection according to a predefined selection ratio. 
Reference-model-based methods have been empirically shown to deliver substantial improvements over full-data baselines, underscoring the promise of token-level selection while also highlighting the need for more practical and flexible approaches.
\subsection{Self-modulated Token Selection}

As discussed in Sec.~\ref{sec:intro}, although reference-model-based token selection demonstrates strong performance, directly adopting a stronger model with the same tokenizer is not always practical, and training a reference model incurs additional time and resource costs. 
We therefore reinterpret token selection as a \emph{self-modulated process}: rather than depending on an external reference, the model leverages its own training trajectory to progressively determine which tokens remain informative and learnable, thereby aligning token selection with its evolving capability.

Specifically, the history model can be defined as the well-pretrained base LLM before SFT, or updated using checkpoints saved during training. Accordingly, we propose \textbf{Retrospective Excess Loss (REL)} as a self-modulated signal that quantifies model progress over its history, serving as a replacement for the EL defined in Sec.~\ref{sec:preliminaries}:
\begin{equation}
\label{eq:rel}
\text{REL}(x_i) \;=\; \mathcal{L}_{\theta_\text{his}}(x_i) - \mathcal{L}_\theta(x_i) 
= \log \frac{\mathbb{P}_\theta(x_i \mid x_{<i})}{\mathbb{P}_{\theta_\text{his}}(x_i \mid x_{<i})}.
\end{equation}
This formulation operationalizes the self-modulated principle: as the current model improves over its historical states, tokens that continue to yield notable gains are adaptively prioritized, whereas noisy or already well-learned ones are down-weighted. In this way, token selection becomes both reference-free and dynamically aligned with the trajectory of model training. Conceptually, if the current model assigns a substantially higher prediction probability to token $x_i$ than the history model, $x_i$ is more likely to represent content that remains learnable at the current stage, rather than tokens already mastered or inherently noisy.

Moreover, \citet{tokencleaning} points out that the capacity of the reference model significantly affects the quality of selected tokens and the final model performance. 
For training on large-scale datasets, however, providing stable long-term guidance becomes particularly important. Unlike reference models, which are often difficult to update continuously, the history model can be easily iterated through training checkpoints, for example via exponential moving average (EMA):
\begin{equation}
\label{eq:ema}
\text{(Optional)} \;\; \theta_{\text{his}}^t = \alpha \, \theta_{\text{his}}^{t-1} + (1-\alpha) \, \theta^t,
\end{equation}
where $\alpha$ controls the smoothing factor. In this way, the history model retains information from past parameters while remaining adaptive, thereby enhancing its potential to provide stable guidance for large-scale, long-horizon training.

Although self-modulated token selection eliminates the need to train or access a separate reference model, the history and current models share identical parameters at the beginning of training, which makes early-stage token selection nearly random. Moreover, similar to prior approaches, self-modulated selection still relies solely on loss information, potentially overlooking tokens that are less favored by loss-based criteria but highly relevant to the task. This limitation motivates us to explore new signals beyond loss that can better capture token importance and task relevance, enabling a more comprehensive token selection strategy.

\subsection{Semantic-aware Token Importance Estimation}

For a well-pretrained LLM, the attention matrix encodes rich semantic information. When answering task-related questions, a token may exhibit persistently high prediction uncertainty (i.e., high loss), yet its attention signal can still play an essential role in predicting subsequent tokens and thus should not be discarded. Motivated by this, we aim to design an attention-based token importance estimation metric that complements loss-based selection with additional semantic signals. Although prior work has proposed estimating token importance using the sum of attentions from other tokens to the target token~\cite{luo2020attention,adnan2024keyformer}, directly applying this approach under the next-token-prediction paradigm with a causal mask introduces additional positional bias, where tokens at different positions receive highly uneven attention. We observe that in the SFT setting, only response tokens need to be selected and used for gradient computation, and all response tokens consistently attend to a fixed-length prompt. This motivates a simple yet effective idea: measuring the total attention each response token assigns to the prompt tokens. Intuitively, since the prompt encodes the task description or instruction to be followed, the extent to which a response token attends to the prompt naturally reflects its task relevance and instruction-following importance, making it a reasonable indicator of token importance.

Specifically, we focus on the attention matrix at a fixed layer $l$ during the forward pass of the model. For each attention head $h \in \{1, \ldots, H\}$ in the decoder, we denote its attention matrix as
\( A^{(h)} \in \mathbb{R}^{L_{\text{seq}} \times L_{\text{seq}}}\), where $A^{(h)}_{i,j}$ represents the attention weight from token $i$ (query) to token $j$ (key), and $L_{\text{seq}} = L_{\text{prompt}} + L_{\text{resp}}$ is the total sequence length. Let $\mathcal{I}_{\text{prompt}} = \{0, 1, \ldots, L_{\text{prompt}} - 1\}$ and $\mathcal{I}_{\text{resp}} = \{L_{\text{prompt}}, \ldots, L_{\text{seq}} - 1\}$ denote the index sets of prompt and response tokens, respectively. To compute the attentions from response tokens to prompt tokens, we extract the following submatrix:
\begin{equation}
A^{(h)}_{\text{resp} \to \text{prompt}} := A^{(h)}[\mathcal{I}{\text{resp}}, \mathcal{I}{\text{prompt}}] \in \mathbb{R}^{L_{\text{resp}} \times L_{\text{prompt}}}.
\end{equation}
By summing over the prompt dimension, we obtain the attention scores of all response tokens in the sequence:
\begin{equation}
\text{AttnScore}^{(h)} = A^{(h)}_{\text{resp} \to \text{prompt}} \cdot \mathbf{1}_{\text{prompt}} \in \mathbb{R}^{L_{\text{resp}}}.
\end{equation}
Subsequently, we average the attention scores obtained from all heads $h \in \{1, \ldots, H\}$, which serves as a token importance estimator complementary to loss information. In summary, the final attention score for a response token $x_i$ is computed as:
\begin{equation}
\label{eq:attnscore}
\text{AttnScore}(x_i) = \frac{1}{H}\sum_{h=1}^{H}\;\mathbf{1}_{\text{prompt}}^{\top}\!\cdot\mathrm{softmax}\!\left(\frac{q_i^{(h)}{\mathbf{K}^{(h)}}^{\top}+\mathbf{M}_i}{\sqrt{d_k}}\right),
\end{equation}
where $\mathbf{M}_i$ denotes the causal mask. An important design question is which layer’s attention matrix should be used to compute the attention scores. In Appendix~\ref{sec:add_results}, we conduct ablation studies comparing early, middle, and deep layers, and conclude that using deeper layers generally yields better results. This finding aligns with prior studies~\cite{aljaafari2024mechanics,zheng2024attention,rocchetti2025instruction}, which suggest that semantic representations become increasingly abstract across layers: shallow layers (closer to the input) primarily capture syntax, local structures, and surface-level patterns such as positional relations, bracket matching, and syntactic cues, while deeper layers (closer to the output) focus more on semantic abstraction, high-level concepts, and task-relevant global information, which are typically more influential for instruction following.

Moreover, prior works in other domains that rely on attention scores for analysis~\cite{chen2024image,ye2025fit} are often incompatible with efficient attention implementations such as FlashAttention~\cite{dao2022flashattention}.  To avoid this, we design a lightweight solution: during the forward pass, we use a hook to store the hidden states of the target layer and then perform a simple recomputation of that layer to retrieve its attention matrix. This design eliminates the need to output full attention matrices during the complete forward pass, thereby making our algorithm fully compatible with efficient attention mechanisms like FlashAttention and ensuring training efficiency.

\subsection{Overall Framework of ssToken}

After obtaining both the retrospective excess loss and the attention score for each token during training, we combine them to perform multi-dimensional token selection. Specifically, due to the Transformer’s computation mechanism and our definition of attention scores, the attention score is naturally bounded within $[0,1]$. For REL, we normalize all response tokens within each sample as
\begin{equation}
\label{eq:norm}
\text{Normalize}(\text{REL}(x_i)) = \frac{\text{REL}(x_i) - \min_j \text{REL}(x_j)}{\max_j \text{REL}(x_j) - \min_j \text{REL}(x_j)},
\end{equation}
which maps the values into the range $[0,1]$. We then introduce a balance coefficient $\gamma \in [0,1]$ to weight the two signals and compute the final score for each token:
\begin{equation}
\text{Score}(x_i) = \gamma \cdot \text{Normalize}(\text{REL}(x_i)) + (1-\gamma) \cdot \text{AttnScore}(x_i).
\end{equation}
Here, $\gamma$ controls the relative contribution of loss-based and attention-based signals in token selection. When $\gamma=0$ or $\gamma=1$, ssToken degenerates into purely attention-based selection or purely loss-based selection, respectively. Our ablation studies in Sec.~\ref{sec:ablation} reveal that appropriate weighting consistently outperforms using either signal alone, and the best performance across benchmarks typically emerges when $\gamma$ takes intermediate values rather than the extremes (i.e., 0 or 1). Based on experimental validation, we set $\gamma=0.5$ as the default choice.

Following~\cite{rho,tokencleaning}, we perform token selection with a fixed ratio $\rho$. Specifically, the loss function in Eq.~\ref{eq:slm} is modified as
\begin{equation}
\mathcal{L}_\theta(\boldsymbol{x}) = - \frac{1}{L_{\text{resp}} \cdot \rho} \sum_{i} I_\rho (x_i) \log \mathbb{P}_\theta(x_i \mid x_{<i}),
\end{equation}
where $L_{\text{resp}} \cdot \rho$ denotes the number of response tokens selected from the sample $\boldsymbol{x}$ according to the top-$\rho$ fraction of $\text{Score}(x_i)$. The indicator function $I_\rho(x_i)$ is defined as
\begin{equation}
I_\rho(x_i) =
\begin{cases}
1, & \text{if } x_i \text{ is among the top-}\rho \text{ tokens ranked by } \text{Score}(x_i), \\
0, & \text{otherwise}.
\end{cases}
\end{equation}
Based on prior empirical evidence~\cite{rho,tokencleaning}, setting $\rho=0.6$ is generally effective for retaining task-relevant tokens while filtering out redundant, noisy, or less informative ones. We further provide a detailed ablation study on $\rho$ in Sec.~\ref{sec:ablation}, offering practical guidance for its selection. The complete algorithmic workflow of ssToken is presented in Appendix~\ref{sec:algo_framework}.

\section{Experiments}
\label{sec:experiments}
\subsection{Protocols}

\textbf{Data Preparation.} Following the setup in~\cite{tokencleaning}, we construct our experimental data pool by sampling 50k examples from a collection of five widely used SFT datasets (300k samples in total): Flan\_v2~\citep{longpre2023flan}, OpenAssistant~\citep{kopf2024openassistant}, Stanford Alpaca~\citep{stanford_alpaca2023}, Dolly~\citep{databricks2023dolly}, and WizardLM~\citep{xu2023wizardlm}. 

\textbf{Base Models.}
We adopt four widely used open-source LLMs of varying sizes, ranging from 3B to 14B parameters: LLaMA-3.2-3B~\cite{dubey2024llama}, LLaMA-3.1-8B~\cite{dubey2024llama}, Qwen-2.5-7B~\cite{hui2024qwen2}, and Qwen-2.5-14B~\cite{hui2024qwen2}. These models are fine-tuned on our curated data pool and compared against different token selection methods. 

\textbf{Baselines.}
For each model, we evaluate several strategies for comparison:
(1) Full-data fine-tuning: training on the entire pool of 50k high-quality samples using supervised fine-tuning.
(2) Uniform Random: selecting a fixed proportion of $\rho$ tokens uniformly at random from the data pool for training.
(3) RHO-1~\cite{rho}: selecting the top-$\rho$ tokens within each sample based on excess loss.
(4) TokenCleaning~\citep{tokencleaning}: including \textit{Fixed-model Cleaning}, which selects the global top-$\rho$ tokens from the entire data pool according to excess loss, and \textit{Self-evolving Cleaning}, which partitions the data pool and iteratively updates the reference model while selecting global top-$\rho$ tokens from each partition. In our experiments, we reproduce the methods based on their released code and report the best performance between the two variants.

Consistent with the setup in~\cite{tokencleaning}, for methods requiring a reference model, the reference is trained on a 10k high-quality subset of the data pool, which is obtained through further filtering with the new powerful sample-level selection method $DS^2$~\cite{pang2024improving}. The token selection ratio $\rho$ is predetermined, and detailed settings as well as related ablation studies are provided in Sec.~\ref{sec:main_results} and Sec.~\ref{sec:ablation}.

\textbf{Evaluation.}
We evaluate our method on a diverse suite of benchmarks that cover different aspects of LLM capabilities, including MMLU~\cite{hendrycks2021measuring}, TriviaQA~\cite{joshi2017triviaqa}, TruthfulQA~\cite{lin2021truthfulqa}, ARC-Easy~\cite{clark2018think}, ARC-Challenge~\cite{clark2018think}, TyDiQA~\cite{clark2020tydi}, Winogrande~\cite{sakaguchi2021winogrande}, HellaSwag~\cite{zellers2019hellaswag}, LogiQA~\cite{liu2020logiqa}, and AGIEval~\cite{zhong2023agieval}. Together, these benchmarks provide a comprehensive assessment of factual knowledge, reasoning ability, and cross-lingual generalization.  Evaluations are conducted using the standard \texttt{lm-eval-harness}\footnote{\url{https://github.com/EleutherAI/lm-evaluation-harness}}~\citep{eval-harness} framework, and additional training and evaluation details can be found in Appendix~\ref{sec:add_details}.

\begin{table*}[t!]

    \centering
     \caption{Performance comparison of different token selection methods across various models and benchmarks. The best scores are highlighted in \textbf{bold}.}
    \resizebox{1\linewidth}{!}{

    \begin{tabular}{l|cccccccccc|c}

\toprule

\textbf{Methods}

& \textbf{TriviaQA} 

& \textbf{TruthfulQA} 

& \textbf{MMLU}

& \textbf{ARC-C}

& \textbf{ARC-E} 

& \textbf{TyDiQA}

& \textbf{Wino} 

& \textbf{HS} 

& \textbf{LogiQA}

& \textbf{AGIEval} 

& \textbf{AVG} \\

\midrule

\multicolumn{12}{c}{\cellcolor{blue!10} \textbf{Base model: LLaMA-3.2-3B}} \\

\midrule

\textsc{Base} 
& $50.85$ & $39.37$ & $56.20$ & $42.15$ & $74.49$ & $19.05$ & $69.06$ & $55.21$ & $29.49$ & $30.04$ & $46.59$
\\

\textsc{Full} 
& $52.50$ & $43.97$ & $\mathbf{57.34}$ & $45.56$ & $76.42$ & $39.48$ & $70.56$ & $55.75$ & $29.80$ & $32.11$ & $50.35$
\\

\textsc{Random} 
& $52.71$ & $43.72$ & $56.84$ & $45.14$ & $76.60$ & $35.66$ & $70.03$ & $55.70$ & $29.95$ & $31.89$ & $49.82$
\\

\textsc{Rho-1} 
& $53.14$ & $44.35$ & $56.78$ & $45.31$ & $76.98$ & $50.47$ & $70.17$ & $56.67$ & $29.80$ & $32.79$ & $51.65$
\\

\textsc{TokenCleaning} 
& $53.09$ & $44.79$ & $57.17$ & $45.22$ & $\mathbf{77.06}$ & $\mathbf{51.40}$ & $69.93$ & $\mathbf{56.69}$ & $30.57$ & $32.81$ & $51.87$
\\

\midrule

\textsc{ssToken} 
& $\mathbf{54.04}$ & $\mathbf{47.42}$ & $57.02$ & $\mathbf{45.82}$ & $76.42$ & $51.04$ & $\mathbf{70.56}$ & $56.19$ & $\mathbf{32.87}$ & $\mathbf{33.97}$ & $\mathbf{\textcolor{purple}{52.50}}$
\\

\toprule

\multicolumn{12}{c}{\cellcolor{blue!10} \textbf{Base model: LLaMA-3.1-8B}} \\

\midrule

\textsc{Base} 
& $61.51$ & $45.22$ & $65.08$ & $51.45$ & $81.57$ & $20.50$ & $73.56$ & $60.01$ & $27.96$ & $35.33$ & $52.22$
\\

\textsc{Full} 
& $65.21$ & $49.03$ & $\mathbf{65.75}$ & $54.69$ & $83.59$ & $54.64$ & $74.74$ & $60.40$ & $29.80$ & $37.07$ & $57.49$
\\

\textsc{Random} 
& $65.23$ & $48.75$ & $65.68$ & $54.27$ & $83.38$ & $54.13$ & $74.59$ & $60.32$ & $30.41$ & $37.41$ & $57.42$
\\

\textsc{Rho-1} 
& $65.39$ & $52.80$ & $65.50$ & $\mathbf{56.23}$ & $\mathbf{83.88}$ & $57.83$ & $76.48$ & $62.46$ & $31.80$ & $37.70$ & $59.01$
\\

\textsc{TokenCleaning} 
& $65.83$ & $53.41$ & $65.28$ & $56.06$ & $83.84$ & $57.13$ & $\mathbf{76.87}$ & $\mathbf{62.55}$ & $32.64$ & $37.81$ & $59.14$
\\

\midrule

\textsc{ssToken}  
& $\mathbf{66.33}$ & $\mathbf{55.55}$ & $65.32$ & $55.29$ & $83.12$ & $\mathbf{59.28}$ & $75.77$ & $61.63$ & $\mathbf{34.10}$ & $\mathbf{37.97}$ & $\mathbf{\textcolor{purple}{59.44}}$
\\

\toprule

\multicolumn{12}{c}{\cellcolor{blue!10} \textbf{Base model: Qwen-2.5-7B}} \\

\midrule

\textsc{Base} 
& $50.03$ & $\mathbf{56.31}$ & $74.18$ & $48.21$ & $80.43$ & $26.67$ & $73.09$ & $59.98$ & $30.72$ & $56.48$ & $55.61$
\\

\textsc{Full} 
& $56.20$ & $50.51$ & $\mathbf{74.22}$ & $50.26$ & $81.19$ & $63.59$ & $73.09$ & $59.85$ & $\mathbf{31.57}$ & $56.85$ & $59.73$
\\

\textsc{Random} 
& $52.08$ & $51.86$ & $74.18$ & $48.21$ & $79.84$ & $63.09$ & $\mathbf{73.32}$ & $60.07$ & $28.33$ & $56.25$ & $58.72$
\\

\textsc{Rho-1} 
& $53.59$ & $46.45$ & $73.91$ & $47.53$ & $79.55$ & $68.73$ & $72.61$ & $59.00$ & $27.65$ & $55.83$ & $58.49$
\\

\textsc{TokenCleaning} 
& $55.57$ & $46.14$ & $73.88$ & $48.72$ & $80.47$ & $68.17$ & $73.24$ & $58.55$ & $27.34$ & $56.17$ & $58.83$
\\

\midrule

\textsc{ssToken} 
& $\mathbf{56.73}$ & $50.75$ & $74.18$ & $\mathbf{50.51}$ & $\mathbf{81.48}$ & $\mathbf{69.74}$ & $72.93$ & $\mathbf{60.30}$ & $31.03$ & $\mathbf{57.18}$ & $\mathbf{\textcolor{purple}{60.48}}$
\\

\toprule

\multicolumn{12}{c}{\cellcolor{blue!10} \textbf{Base model: Qwen-2.5-14B}} \\

\midrule

\textsc{Base} 
& $59.42$ & $\mathbf{58.45}$ & $79.80$ & $55.80$ & $82.32$ & $28.89$ & $75.06$ & $63.36$ & $38.10$ & $64.27$ & $60.55$
\\

\textsc{Full} 
& $66.36$ & $53.77$ & $79.75$ & $56.40$ & $82.83$ & $64.59$ & $74.98$ & $\mathbf{63.37}$ & $36.87$ & $66.06$ & $64.50$
\\

\textsc{Random} 
& $63.91$ & $53.86$ & $79.92$ & $56.31$ & $82.87$ & $65.32$ & $75.53$ & $63.30$ & $39.02$ & $66.19$ & $64.62$
\\

\textsc{Rho-1} 
& $65.28$ & $52.14$ & $79.62$ & $\mathbf{56.57}$ & $\mathbf{83.71}$ & $71.21$ & $75.85$ & $63.03$ & $35.33$ & $65.23$ & $64.80$
\\

\textsc{TokenCleaning} 
& $66.42$ & $51.87$ & $79.66$ & $55.97$ & $83.08$ & $69.37$ & $\mathbf{76.56}$ & $62.33$ & $38.56$ & $64.86$ & $64.87$
\\

\midrule

\textsc{ssToken} 
& $\mathbf{66.55}$ & $52.96$ & $\mathbf{79.97}$ & $56.22$ & $82.49$ & $\mathbf{71.94}$ & $76.24$ & $62.98$ & $\mathbf{42.70}$ & $\mathbf{66.30}$ & $\mathbf{\textcolor{purple}{65.84}}$
\\

\bottomrule

\end{tabular}

    }

    \label{tab:main_results}

\end{table*}

\subsection{Main Results}
\label{sec:main_results}

We conduct experiments on four different base models with five token selection strategies (including our proposed ssToken) and evaluate their performance across ten general-domain benchmarks. The detailed results are presented in Table~\ref{tab:main_results}. The balance coefficient $\gamma$ is set to 0.5 across all model sizes. The token selection ratio $\rho$ is set to 0.6 for LLaMA-3.2-3B, LLaMA-3.1-8B, and Qwen-2.5-7B, and 0.8 for Qwen-2.5-14B. For the same base model, all token selection methods are evaluated under the same ratio. Detailed ablation studies can be found in Sec.~\ref{sec:ablation}.

As shown in Table~\ref{tab:main_results}, ssToken achieves the best average performance across all four base models compared to other existing token selection methods. Relative to the full-data fine-tuning baseline, ssToken improves the average performance by \textbf{4.3\%}, \textbf{3.4\%}, \textbf{1.3\%}, and \textbf{2.1\%} on LLaMA-3.2-3B, LLaMA-3.1-8B, Qwen-2.5-7B, and Qwen-2.5-14B, respectively. We also observe that although prior token selection methods RHO-1 and TokenCleaning show strong performance on LLaMA-3.2-3B and LLaMA-3.1-8B, their performance on the Qwen family is only marginally better than—or sometimes even worse than—that of full-data fine-tuning. In contrast, our proposed ssToken consistently delivers strong and stable gains across different model sizes and families, further demonstrating its effectiveness and generality.

Furthermore, we find that the impact of token selection varies across different benchmarks. For knowledge-intensive tasks such as MMLU, ARC-Challenge, and ARC-Easy, which heavily rely on pretraining, token selection during SFT does not lead to notable performance improvements. In contrast, for benchmarks like TyDiQA, TriviaQA, and AGIEval that target downstream tasks or require stronger instruction-following ability, SFT typically yields substantial gains for the base models, and token selection often provides further improvements. Notably, ssToken outperforms previous token selection methods on nearly all QA tasks across the four models. Since QA tasks generally demand stronger instruction-following capability, we attribute this advantage to the attention-based component of ssToken, which complements loss information with rich semantic cues and effectively guides the selection of instruction-relevant tokens.

\begin{wrapfigure}{r}{0.45\linewidth}
    \centering
    \includegraphics[width=\linewidth]{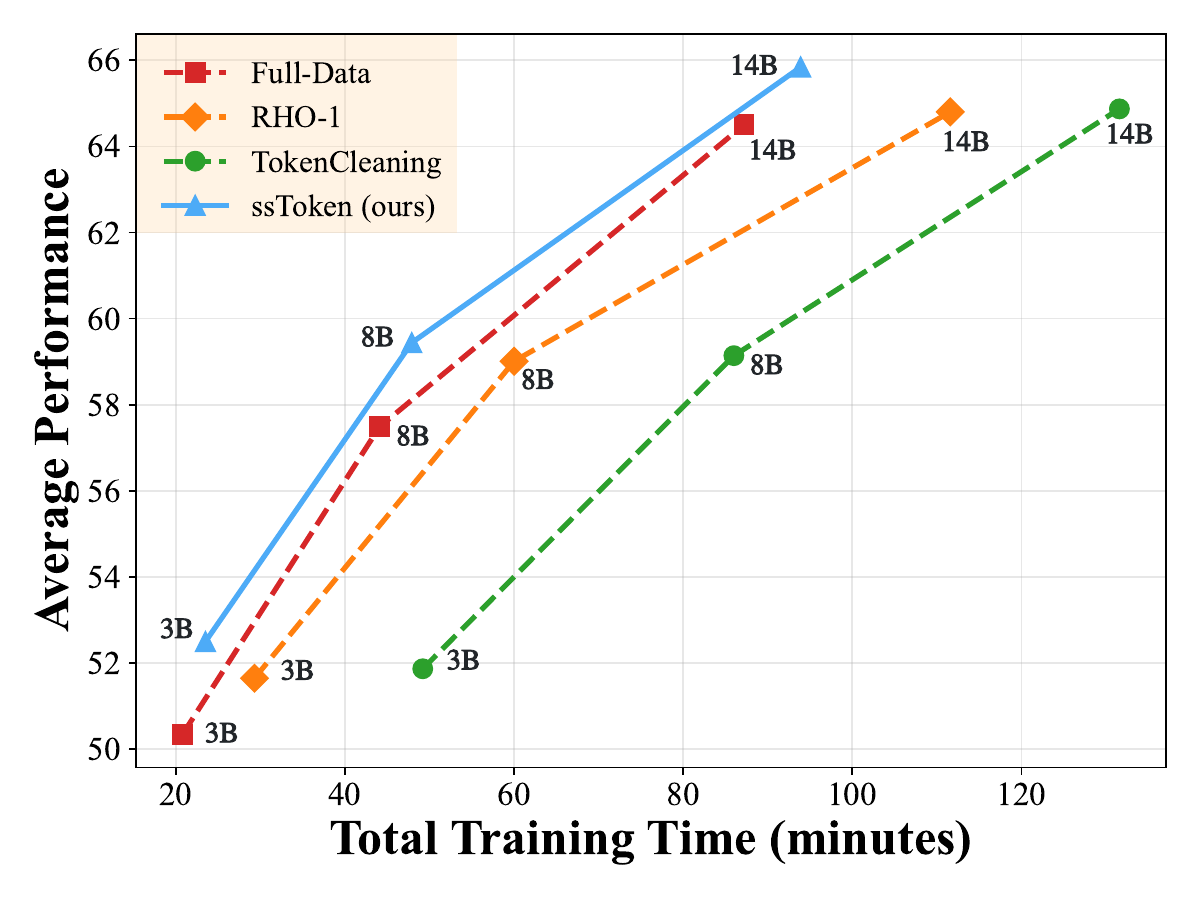}
    \caption{Average performance vs. total training time across different methods.}
    \label{fig:performance_time} 
\end{wrapfigure}
\textbf{Performance vs. Training Time.}
In Fig.~\ref{fig:performance_time}, we further compare the performance and total training time (including model loading and reference model training) of different token selection methods against the full-data fine-tuning baseline across various model sizes.
It is worth noting that existing token selection approaches are primarily designed to enhance performance rather than reduce training cost. Specifically, while the filtered-out tokens are masked during loss computation, they still participate in the forward pass, and thus these methods do not yield reductions in training time compared to full-data fine-tuning.
As shown in the figure, while RHO-1 and TokenCleaning achieve notable performance gains over the full-data baseline, their reliance on a reference model incurs substantial additional time and resource overhead. In contrast, ssToken does not need to train an extra reference model, and its lightweight design for obtaining attention matrices introduces marginal computational cost. Consequently, compared with the full-data baseline, ssToken delivers significant performance improvements with only a marginal increase in training time, thereby maintaining training efficiency.

\subsection{Ablation Studies}
\label{sec:ablation}

Our proposed ssToken involves a key hyperparameter, the balance coefficient $\gamma$, which controls the relative contributions of loss-based and attention-based information. In addition, similar to previous token selection methods, ssToken also requires a predefined ratio $\rho$ for selecting tokens. We therefore conduct experiments on these two hyperparameters across models of different sizes, aiming to investigate their general effects on model performance and to provide practical guidance for their 
appropriate settings.

\textbf{Ablation studies on the balance coefficient $\gamma$.}
We report the average fine-tuning performance across different model sizes with $\gamma \in \{0, 0.25, 0.5, 0.75, 1\}$. The results are visualized in Fig.~\ref{fig:ablation_gamma}, while detailed results across all benchmarks are provided in Appendix~\ref{sec:add_results}. In these experiments, only $\gamma$ is varied, with all other settings (including the token selection ratio $\rho$) kept fixed.

As shown in the figure, compared with the full-data fine-tuning baseline, both self-modulated token selection ($\gamma=1$) and semantic-aware token selection ($\gamma=0$) achieve superior performance, while combining the two signals in proper proportions yields synergistic effects and further improvements. This validates our hypothesis that loss-based and attention-based information capture complementary features: the latter provides additional semantic cues beyond what is reflected in loss values.

From the results, we observe that $\gamma=0.5$ achieves the best performance for models with 8B and 14B parameters, and ranks second for the 3B model (LLaMA-3.2-3B), slightly behind $\gamma=0.75$. A closer look at the detailed results in Appendix~\ref{sec:ablation_details} shows that the advantage of $\gamma=0.75$ on LLaMA-3.2-3B mainly comes from its strong performance on TyDiQA and TruthfulQA. In contrast, although $\gamma=0.5$ is marginally weaker than $\gamma=0.75$ on LLaMA-3.2-3B, it achieves more balanced performance across benchmarks. Considering these observations, together with its optimal results on larger models (8B and 14B), we recommend setting $\gamma=0.5$ as the default choice.

\begin{figure}[t]
  \centering
  \begin{subfigure}[t]{0.45\linewidth}
    \centering
    \includegraphics[width=\linewidth]{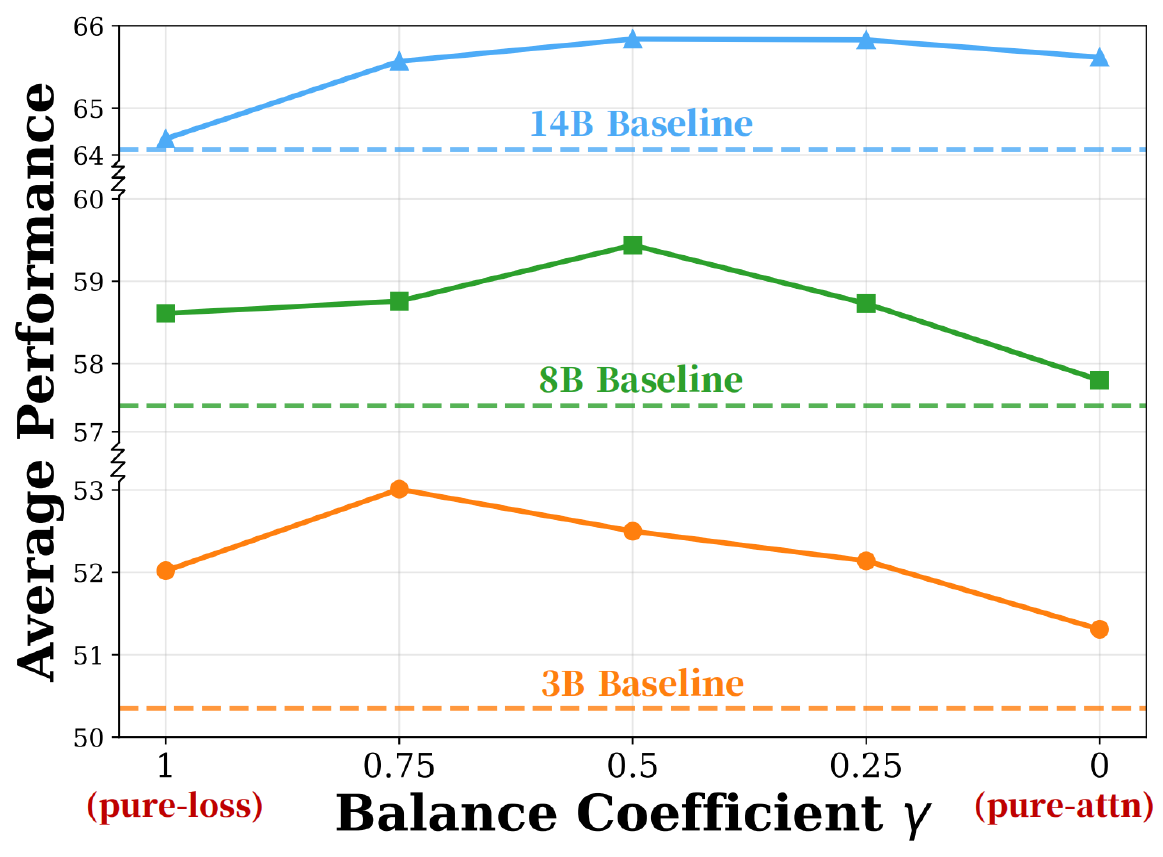}
    \subcaption{Ablation study on the balance coefficient $\gamma$}
    \label{fig:ablation_gamma}
  \end{subfigure}
  \hfill
  \begin{subfigure}[t]{0.45\linewidth}
    \centering
    \includegraphics[width=\linewidth]{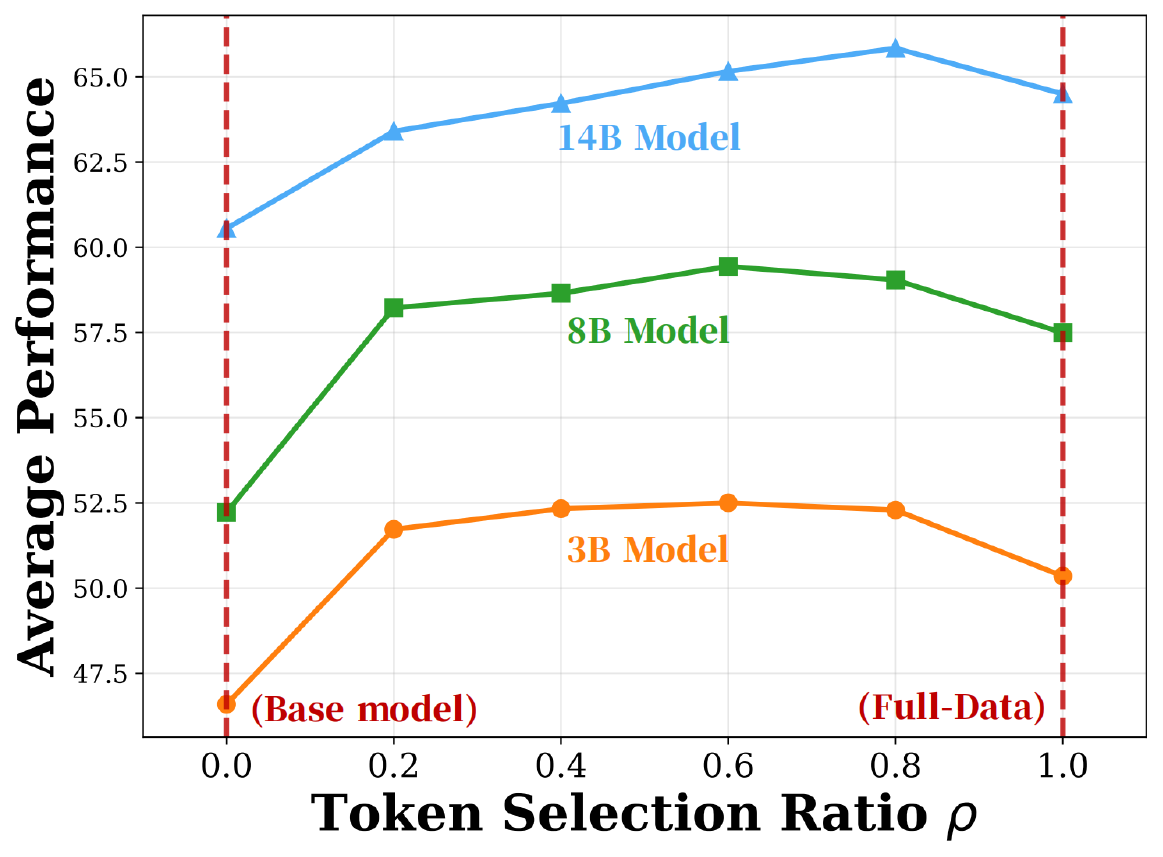}
    \subcaption{Ablation study on token selection ratio $\rho$}
    \label{fig:ablation_rho}
  \end{subfigure}
  \vspace{-2pt}
\caption{Ablation studies on $\gamma$ and $\rho$. (a) Varying the balance coefficient $\gamma$, with dashed lines indicating the full-data fine-tuning baselines for different model sizes. (b) Varying the token selection ratio $\rho$. Both plots report average performance across model sizes (3B, 8B, 14B).}
  \label{fig:ablation_pair}
  \vspace{-10pt}
\end{figure}
  
\textbf{Ablation studies on the token selection ratio $\rho$.}
We report the average fine-tuning performance across different model sizes with $\rho \in \{0, 0.2, 0.4, 0.6, 0.8, 1\}$. The results are visualized in Fig.~\ref{fig:ablation_rho}, and detailed results across all benchmarks are provided in Appendix~\ref{sec:add_results}. In these experiments, only $\rho$ is varied, while all other settings are kept fixed.

As shown in the figure, although the SFT data we use has already been curated through strong data cleaning pipelines, applying token-level selection with an appropriate ratio still yields further improvements over full-data fine-tuning across different model sizes. This provides compelling evidence for the promise of token-level selection. For LLaMA-3.2-3B and LLaMA-3.1-8B, the best results are achieved at $\rho=0.6$, consistent with the ablation findings reported in~\cite{tokencleaning}. However, for Qwen-2.5-14B, we observe that not only ssToken but also other token selection methods (Random, RHO-1, TokenCleaning) reach their peak performance at $\rho=0.8$ (detailed results on this comparison are presented in Appendix~\ref{sec:ablation_details}). We hypothesize that this phenomenon may stem from differences in pretraining data distributions, which affect the capability boundaries of different model families; in addition, larger models may better capture more challenging task-related patterns that smaller models fail to recognize. In summary, the choice of $\rho$ should be guided by both model capacity and data quality. In line with prior works~\cite{rho,tokencleaning}, our ablations suggest that $\rho=0.6$ is generally a robust choice, while slightly larger values may yield further improvements in certain settings.

\section{Conclusion, Limitations and Future Work}
In this paper, we introduced \textbf{ssToken}, a self-modulated and semantic-aware token selection framework for LLM supervised fine-tuning. By formulating token selection as a self-modulated process, ssToken leverages retrospective excess loss over historical model states to adaptively prioritize tokens that remain informative and learnable, thereby eliminating the need for training an additional reference model. Moreover, by leveraging an attention-based score tailored for the SFT setting, ssToken provides a semantic complement to loss-based information. Experiments demonstrate that both self-modulated selection and semantic-aware selection alone outperform full-data fine-tuning, while combining the two signals yields synergistic gains and further improvements. Extensive results across different model families and scales show that ssToken consistently surpasses both full-data baselines and prior token selection methods, achieving performance improvements while maintaining training efficiency.

\textbf{Limitations and Future Work.}
Similar to previous token selection methods, ssToken requires a predefined token selection ratio $\rho$ as the threshold for filtering tokens. However, as shown in the ablation studies in Sec.~\ref{sec:ablation}, the optimal value of $\rho$ is not universal but rather depends on both the capability of the base model and the quality of the training data. This reliance on manual tuning introduces additional overhead and may limit the generalizability of the method across different model families or domains. Instead of determining $\rho$ through extensive ablation studies, a more promising research direction is to integrate its optimization into the algorithm itself, enabling $\rho$ to be adaptively adjusted during training according to model progress, capacity, and data quality. We leave the development of such adaptive mechanisms for future work.

\clearpage

\bibliography{iclr2026_conference}
\bibliographystyle{iclr2026_conference}

\clearpage
\appendix

\section{Experiment Details}
\label{sec:add_details}
\subsection{Data Pool}\label{sec:apx_data_pool}
Following~\cite{tokencleaning}, we construct our data pool from five commonly adopted SFT datasets, which are either manually curated or generated with the assistance of powerful LLMs. These datasets vary significantly in terms of prompt style, annotation quality, average length, and task focus, thereby ensuring that the resulting pool is both representative and diverse.  
To maintain consistency across sources, we standardize all datasets using the template introduced by \citet{wang2023far}, which explicitly marks conversational roles with \texttt{<|User|>} and \texttt{<|Assistant|>} tags.

\subsection{Evaluation Benchmarks}\label{sec:appendix_evalution_details}
We evaluate models on a wide range of benchmarks that capture different aspects of LLM capability, again following~\cite{tokencleaning}. For datasets such as MMLU, LogiQA, ARC-C, ARC-E, Winogrande, AGIEval and HellaSwag, we report accuracy as the evaluation metric. In particular, for TruthfulQA, which is formulated as a multiple-choice task, we adopt the MC2 metric that scores only the answer assigned the highest probability by the model. For TriviaQA, we use the \textit{exact\_match} metric. For TyDiQA, we use the 1-shot F1 score. All evaluations are performed using the official \texttt{lm-evaluation-harness} toolkit, with the default setting of evaluating on the full benchmark sample set.

\subsection{Training Details}
Following the setup in~\cite{tokencleaning}, we adopt the LoRA technique~\cite{hu2022lora} with a rank of 64 and a scaling factor of 16. Training is conducted for one epoch with a total batch size of 48 and a learning rate of $1\times10^{-4}$. The maximum input sequence length is set to 2048 tokens. All experiments are conducted on eight NVIDIA H200 GPUs. For the training time reported in Fig.~\ref{fig:performance_time}, we present the complete runtime of each token selection method, including both model loading and the training of the reference model.
Due to the relatively small size of the SFT data pool, we observe that enabling the optional EMA-based updating of the history model does not yield noticeable performance gains, while model merging introduces additional computational overhead. Therefore, in all experiments we use a fixed base model as the history model. We hypothesize that adaptive updating of the history model may demonstrate greater potential in large-horizon training scenarios. 

\section{Additional Experimental Results}
\label{sec:add_results}
\subsection{Ablation Study on Attention Layer Depth}
In our semantic-aware token importance estimation, we require the attention matrix from a specific layer of the forward pass, and the choice of layer can influence the effectiveness of token selection. To investigate this, we conduct experiments on LLaMA-3.2-3B by extracting attention matrices from shallow, middle, and deep layers, respectively. The results are reported in Table~\ref{tab:layer}, with the balancing coefficient $\gamma$ fixed at 0.5 and the token selection ratio $\rho$ set to 0.6. 

\begin{table*}[h!]
    \centering
    \caption{Ablation study on attention layer depth for ssToken on LLaMA-3.2-3B. Results are reported on ten benchmarks, with the best score in each column highlighted in \textbf{bold}.}
    \resizebox{1\linewidth}{!}{
    \begin{tabular}{l|cccccccccc|c}
    \toprule
    \textbf{Methods} 
    & \textbf{TriviaQA} 
    & \textbf{TruthfulQA} 
    & \textbf{MMLU} 
    & \textbf{ARC-C} 
    & \textbf{ARC-E} 
    & \textbf{TyDiQA} 
    & \textbf{Wino} 
    & \textbf{HS} 
    & \textbf{LogiQA} 
    & \textbf{AGIEval} 
    & \textbf{AVG} \\
    \midrule
    \multicolumn{12}{c}{\cellcolor{blue!10} \textbf{Base model: LLaMA-3.2-3B}} \\
    \midrule
    \textsc{Base} 
    & $50.85$ & $39.37$ & $56.20$ & $42.15$ & $74.49$ & $19.05$ & $69.06$ & $55.21$ & $29.49$ & $30.04$ & $46.59$ \\
    \textsc{Full-data} 
    & $52.50$ & $43.97$ & $57.34$ & $45.56$ & $76.42$ & $39.48$ & $70.56$ & $55.75$ & $29.80$ & $32.11$ & $50.35$ \\
    \midrule
    \textsc{Shallow} 
    & $54.36$ & $45.12$ & $56.23$ & $45.48$ & $\mathbf{77.48}$ & $48.52$ & $70.09$ & $53.76$ & $31.80$ & $33.43$ & $51.63$ \\
    \textsc{Medium} 
    & $\mathbf{54.45}$ & $47.04$ & $\mathbf{57.43}$ & $45.73$ & $77.23$ & $49.20$ & $69.77$ & $55.40$ & $29.95$ & $32.96$ & $51.92$ \\
    \textsc{Deep} 
    & $54.04$ & $\mathbf{47.42}$ & $57.02$ & $\mathbf{45.82}$ & $76.42$ & $\mathbf{51.04}$ & $\mathbf{70.17}$ & $\mathbf{56.19}$ & $\mathbf{32.87}$ & $\mathbf{33.97}$ & $\mathbf{52.50}$ \\
    \bottomrule
    \end{tabular}}
    \label{tab:layer}
\end{table*}
As shown in Table~\ref{tab:layer}, the choice of attention layer has a clear impact on token selection performance. As the layer depth increases, the average performance of the fine-tuned model consistently improves.
This trend aligns with prior findings on layerwise representations in Transformer-based LLMs~\cite{aljaafari2024mechanics,zheng2024attention,rocchetti2025instruction}: shallow layers primarily encode local syntactic or positional information, while deeper layers capture more abstract semantic features and task-relevant dependencies. Since our semantic-aware metric measures the importance of response tokens relative to the prompt, deeper attention distributions provide stronger semantic cues, thereby improving the identification of instruction-relevant tokens.

\subsection{Detailed Ablation on $\gamma$ and $\rho$}
\label{sec:ablation_details}
In Sec.~\ref{sec:ablation} of the main text, we presented an ablation study on the hyperparameters $\gamma$ and $\rho$, with the final results briefly summarized in Fig.~\ref{fig:ablation_pair}. In this section, we provide the complete experimental results of ssToken with respect to $\gamma$ and $\rho$, reported in Table.~\ref{tab:sstoken_ratios_all} and Table.~\ref{tab:ablation_rho}. In addition, Table.~\ref{tab:qwen14b_rho_compare_methods} compares the performance of different token selection methods on Qwen-2.5-14B under two settings, $\rho=0.6$ and $\rho=0.8$.

\begin{table*}[b]
    \centering
    \caption{Performance of \textsc{ssToken} with different balance coefficient $\gamma$ across different base models and various benchmarks.}
    \resizebox{1\linewidth}{!}{
    \begin{tabular}{l|cccccccccc|c}
    \toprule
    \textbf{Methods}
    & \textbf{TriviaQA} 
    & \textbf{TruthfulQA} 
    & \textbf{MMLU}
    & \textbf{ARC-C}
    & \textbf{ARC-E} 
    & \textbf{TyDiQA}
    & \textbf{Wino} 
    & \textbf{HS} 
    & \textbf{LogiQA}
    & \textbf{AGIEval} 
    & \textbf{AVG} \\
    \midrule
    \multicolumn{12}{c}{\cellcolor{blue!10} \textbf{Base model: LLaMA-3.2-3B}} \\
    \midrule
    \textsc{Base} 
    & $50.85$ & $39.37$ & $56.20$ & $42.15$ & $74.49$ & $19.05$ & $69.06$ & $55.21$ & $29.49$ & $30.04$ & $46.59$ \\
    \textsc{Full} 
    & $52.50$ & $43.97$ & $57.34$ & $45.56$ & $76.42$ & $39.48$ & $70.56$ & $55.75$ & $29.80$ & $32.11$ & $50.35$ \\
    \textsc{ssToken ($\gamma=1.0$)} 
    & $51.04$ & $48.34$ & $56.34$ & $43.26$ & $74.66$ & $56.97$ & $68.51$ & $51.78$ & $34.87$ & $34.39$ & $52.02$ \\
    \textsc{ssToken ($\gamma=0.75$)} 
    & $52.64$ & $51.86$ & $56.72$ & $44.97$ & $75.97$ & $55.64$ & $70.17$ & $54.50$ & $33.03$ & $34.60$ & $53.01$ \\
    \textsc{ssToken ($\gamma=0.5$)} 
    & $54.04$ & $47.42$ & $57.02$ & $45.82$ & $76.42$ & $51.04$ & $70.17$ & $56.19$ & $32.87$ & $33.97$ & $52.50$ \\
    \textsc{ssToken ($\gamma=0.25$)} 
    & $54.46$ & $46.62$ & $57.31$ & $45.82$ & $77.36$ & $49.11$ & $70.72$ & $55.75$ & $31.34$ & $32.87$ & $52.14$ \\
    \textsc{ssToken ($\gamma=0.0$)} 
    & $53.32$ & $45.35$ & $56.93$ & $45.48$ & $77.10$ & $47.27$ & $70.32$ & $55.75$ & $29.65$ & $31.93$ & $51.31$ \\
    \midrule
    \multicolumn{12}{c}{\cellcolor{blue!10} \textbf{Base model: LLaMA-3.1-8B}} \\
    \midrule
    \textsc{Base} 
    & $61.51$ & $45.22$ & $65.08$ & $51.45$ & $81.57$ & $20.50$ & $73.56$ & $60.01$ & $27.96$ & $35.33$ & $52.22$ \\
    \textsc{Full} 
    & $65.21$ & $49.03$ & $65.75$ & $54.69$ & $83.59$ & $54.64$ & $74.74$ & $60.40$ & $29.80$ & $37.07$ & $57.49$ \\
    \textsc{ssToken ($\gamma=1.0$)} 
    & $65.50$ & $56.51$ & $64.70$ & $53.33$ & $82.07$ & $60.60$ & $74.90$ & $56.78$ & $32.87$ & $38.81$ & $58.61$ \\
    \textsc{ssToken ($\gamma=0.75$)} 
    & $65.87$ & $54.24$ & $65.63$ & $53.75$ & $82.07$ & $59.23$ & $75.69$ & $59.93$ & $32.87$ & $38.28$ & $58.76$ \\
    \textsc{ssToken ($\gamma=0.5$)} 
    & $66.33$ & $55.55$ & $65.32$ & $55.29$ & $83.12$ & $59.28$ & $75.77$ & $61.63$ & $34.10$ & $37.97$ & $59.44$ \\
    \textsc{ssToken ($\gamma=0.25$)} 
    & $65.54$ & $53.57$ & $65.63$ & $55.29$ & $83.25$ & $58.40$ & $75.45$ & $60.24$ & $32.72$ & $37.19$ & $58.73$ \\
    \textsc{ssToken ($\gamma=0.0$)} 
    & $65.72$ & $48.87$ & $65.83$ & $54.86$ & $83.63$ & $54.67$ & $75.22$ & $60.60$ & $30.88$ & $37.74$ & $57.80$ \\
    \midrule
    \multicolumn{12}{c}{\cellcolor{blue!10} \textbf{Base model: Qwen-2.5-14B}} \\
    \midrule
    \textsc{Base} 
    & $59.42$ & $58.45$ & $79.80$ & $55.80$ & $82.32$ & $28.89$ & $75.06$ & $63.36$ & $38.10$ & $64.27$ & $60.55$ \\
    \textsc{Full} 
    & $66.36$ & $53.77$ & $79.75$ & $56.40$ & $82.83$ & $64.59$ & $74.98$ & $63.37$ & $36.87$ & $66.06$ & $64.50$ \\
    \textsc{ssToken ($\gamma=1.0$)} 
    & $64.41$ & $52.62$ & $79.69$ & $55.38$ & $82.45$ & $69.31$ & $75.61$ & $63.01$ & $37.64$ & $66.18$ & $64.63$ \\
    \textsc{ssToken ($\gamma=0.75$)} 
    & $66.64$ & $52.15$ & $77.93$ & $56.40$ & $83.21$ & $70.91$ & $75.69$ & $63.10$ & $43.32$ & $66.30$ & $65.57$ \\
    \textsc{ssToken ($\gamma=0.5$)} 
    & $66.55$ & $52.96$ & $79.97$ & $56.22$ & $82.49$ & $71.94$ & $76.24$ & $62.98$ & $42.70$ & $66.30$ & $65.84$ \\
    \textsc{ssToken ($\gamma=0.25$)} 
    & $67.04$ & $53.17$ & $79.90$ & $55.55$ & $82.32$ & $72.30$ & $75.85$ & $62.91$ & $43.16$ & $66.10$ & $65.83$ \\
    \textsc{ssToken ($\gamma=0.0$)} 
    & $66.16$ & $54.06$ & $79.92$ & $56.14$ & $83.12$ & $73.53$ & $76.01$ & $63.22$ & $37.94$ & $66.11$ & $65.62$ \\
    \bottomrule
    \end{tabular}}
    \label{tab:sstoken_ratios_all}
\end{table*}

\begin{table*}[t!]
    \centering
    \caption{Performance of ssToken under different token selection ratios $\rho$ across different base models and various benchmarks.}
    \resizebox{1\linewidth}{!}{
    \begin{tabular}{l|cccccccccc|c}
\toprule
\textbf{Methods}
& \textbf{TriviaQA}
& \textbf{TruthfulQA}
& \textbf{MMLU}
& \textbf{ARC-C}
& \textbf{ARC-E}
& \textbf{TyDiQA}
& \textbf{Wino}
& \textbf{HS}
& \textbf{LogiQA}
& \textbf{AGIEval}
& \textbf{AVG} \\
\midrule

\multicolumn{12}{c}{\cellcolor{blue!10} \textbf{Base model: LLaMA-3.2-3B}} \\
\midrule
\textsc{Base ($\rho=0$)} & $50.85$ & $39.37$ & $56.20$ & $42.15$ & $74.49$ & $19.05$ & $69.06$ & $55.21$ & $29.49$ & $30.04$ & $46.59$ \\
\textsc{Full ($\rho=1$)} & $52.50$ & $43.97$ & $57.34$ & $45.56$ & $76.42$ & $39.48$ & $70.56$ & $55.75$ & $29.80$ & $32.11$ & $50.35$ \\
\textsc{ssToken ($\rho=0.2$)} & $51.70$ & $53.82$ & $55.63$ & $44.20$ & $73.65$ & $47.50$ & $70.48$ & $54.32$ & $33.03$ & $32.91$ & $51.72$ \\
\textsc{ssToken ($\rho=0.4$)} & $51.79$ & $51.62$ & $56.22$ & $45.65$ & $75.55$ & $50.70$ & $70.80$ & $55.73$ & $31.80$ & $33.47$ & $52.33$ \\
\textsc{ssToken ($\rho=0.6$)} & $54.04$ & $47.42$ & $57.02$ & $45.82$ & $76.42$ & $51.04$ & $70.17$ & $56.19$ & $32.87$ & $33.97$ & $52.50$ \\
\textsc{ssToken ($\rho=0.8$)} & $53.99$ & $45.72$ & $57.49$ & $45.65$ & $77.57$ & $50.10$ & $70.24$ & $56.26$ & $31.95$ & $33.88$ & $52.29$ \\

\toprule
\multicolumn{12}{c}{\cellcolor{blue!10} \textbf{Base model: LLaMA-3.1-8B}} \\
\midrule
\textsc{Base ($\rho=0$)} & $61.51$ & $45.22$ & $65.08$ & $51.45$ & $81.57$ & $20.50$ & $73.56$ & $60.01$ & $27.96$ & $35.33$ & $52.22$ \\
\textsc{Full ($\rho=1$)} & $65.21$ & $49.03$ & $65.75$ & $54.69$ & $83.59$ & $54.64$ & $74.74$ & $60.40$ & $29.80$ & $37.07$ & $57.49$ \\
\textsc{ssToken ($\rho=0.2$)} & $66.60$ & $57.28$ & $65.62$ & $52.47$ & $80.93$ & $56.45$ & $74.98$ & $56.48$ & $32.41$ & $38.95$ & $58.22$ \\
\textsc{ssToken ($\rho=0.4$)} & $66.70$ & $55.42$ & $64.98$ & $53.41$ & $82.15$ & $57.98$ & $75.14$ & $57.38$ & $36.25$ & $37.11$ & $58.65$ \\
\textsc{ssToken ($\rho=0.6$)} & $66.33$ & $55.55$ & $65.32$ & $55.29$ & $83.12$ & $59.28$ & $75.77$ & $61.63$ & $34.10$ & $37.97$ & $59.44$ \\
\textsc{ssToken ($\rho=0.8$)} & $65.60$ & $52.13$ & $65.22$ & $54.95$ & $83.38$ & $61.69$ & $75.22$ & $61.77$ & $33.18$ & $37.25$ & $59.04$ \\

\toprule
\multicolumn{12}{c}{\cellcolor{blue!10} \textbf{Base model: Qwen-2.5-14B}} \\
\midrule
\textsc{Base ($\rho=0$)} & $59.42$ & $58.45$ & $79.80$ & $55.80$ & $82.32$ & $28.89$ & $75.06$ & $63.36$ & $38.10$ & $64.27$ & $60.55$ \\
\textsc{Full ($\rho=1$)} & $66.36$ & $53.77$ & $79.75$ & $56.40$ & $82.83$ & $64.59$ & $74.98$ & $63.37$ & $36.87$ & $66.06$ & $64.50$ \\
\textsc{ssToken ($\rho=0.2$)} & $68.71$ & $53.88$ & $78.57$ & $56.66$ & $82.49$ & $61.34$ & $75.61$ & $59.41$ & $33.03$ & $64.33$ & $63.40$ \\
\textsc{ssToken ($\rho=0.4$)} & $64.31$ & $52.71$ & $79.68$ & $54.61$ & $81.73$ & $70.33$ & $75.61$ & $61.76$ & $36.41$ & $65.05$ & $64.22$ \\
\textsc{ssToken ($\rho=0.6$)} & $64.07$ & $52.90$ & $79.88$ & $56.57$ & $83.63$ & $71.94$ & $75.85$ & $62.62$ & $38.71$ & $65.46$ & $65.16$ \\
\textsc{ssToken ($\rho=0.8$)} & $66.55$ & $52.96$ & $79.97$ & $56.22$ & $82.49$ & $71.94$ & $76.24$ & $62.98$ & $42.70$ & $66.30$ & $65.84$ \\

\bottomrule
\end{tabular}}
\label{tab:ablation_rho}
\end{table*}

\begin{table*}[b]
    \centering
    \caption{Comparison of token selection methods on \textbf{Qwen-2.5-14B} at two ratios, $\rho\!=\!0.6$ and $\rho\!=\!0.8$.}
    \resizebox{1\linewidth}{!}{
    \begin{tabular}{l|cccccccccc|c}
    \toprule
    \textbf{Methods}
    & \textbf{TriviaQA}
    & \textbf{TruthfulQA}
    & \textbf{MMLU}
    & \textbf{ARC-C}
    & \textbf{ARC-E}
    & \textbf{TyDiQA}
    & \textbf{Wino}
    & \textbf{HS}
    & \textbf{LogiQA}
    & \textbf{AGIEval}
    & \textbf{AVG} \\
    \midrule
    \multicolumn{12}{c}{\cellcolor{blue!10} \textbf{Base model: Qwen-2.5-14B}} \\
    \midrule
    \textsc{Base} 
    & $59.42$ & $58.45$ & $79.80$ & $55.80$ & $82.32$ & $28.89$ & $75.06$ & $63.36$ & $38.10$ & $64.27$ & $60.55$ \\
    \textsc{Full} 
    & $66.36$ & $53.77$ & $79.75$ & $56.40$ & $82.83$ & $64.59$ & $74.98$ & $63.37$ & $36.87$ & $66.06$ & $64.50$ \\
    \midrule
    \textsc{Random ($\rho=0.6$)} 
    & $65.91$ & $53.57$ & $79.85$ & $56.31$ & $83.50$ & $63.32$ & $75.22$ & $63.33$ & $33.33$ & $66.10$ & $64.04$ \\
    \rowcolor{blue!10}\textsc{Random ($\rho=0.8$)} 
    & $63.91$ & $53.86$ & $79.92$ & $56.31$ & $82.87$ & $65.32$ & $75.53$ & $63.30$ & $39.02$ & $66.19$ & $64.62$ \\
    \midrule
    \textsc{RHO-1 ($\rho=0.6$)} 
    & $62.16$ & $51.71$ & $79.55$ & $56.06$ & $82.91$ & $69.50$ & $76.48$ & $62.65$ & $35.33$ & $65.07$ & $64.14$ \\
    \rowcolor{blue!10}\textsc{RHO-1 ($\rho=0.8$)} 
    & $65.28$ & $52.14$ & $79.62$ & $56.57$ & $83.71$ & $71.21$ & $75.85$ & $63.03$ & $35.33$ & $65.23$ & $64.80$ \\
    \midrule
    \textsc{TokenCleaning ($\rho=0.6$)} 
    & $65.58$ & $51.35$ & $79.70$ & $55.89$ & $82.49$ & $70.02$ & $76.56$ & $62.34$ & $38.56$ & $65.17$ & $64.77$ \\
    \rowcolor{blue!10}\textsc{TokenCleaning ($\rho=0.8$)} 
    & $66.42$ & $51.87$ & $79.66$ & $55.97$ & $83.08$ & $69.37$ & $76.56$ & $62.33$ & $38.56$ & $64.86$ & $64.87$ \\
    \midrule
    \textsc{ssToken ($\rho=0.6$)} 
    & $64.07$ & $52.90$ & $79.88$ & $56.57$ & $83.63$ & $71.94$ & $75.85$ & $62.62$ & $38.71$ & $65.46$ & $65.16$ \\
    \rowcolor{blue!10}\textsc{ssToken ($\rho=0.8$)} 
    & $66.55$ & $52.96$ & $79.97$ & $56.22$ & $82.49$ & $71.94$ & $76.24$ & $62.98$ & $42.70$ & $66.30$ & $65.84$ \\
    \bottomrule
    \end{tabular}}
    \label{tab:qwen14b_rho_compare_methods}
\end{table*}

\section{Algorithmic Framework of ssToken}
\label{sec:algo_framework}
In this section, we summarize the overall algorithmic framework of ssToken in Algorithm~\ref{alg}, highlighting its two key components: self-modulated token selection and semantic-aware token importance estimation.

\begin{algorithm}[h]
\caption{ssToken: Self-modulated and Semantic-aware Token Selection}
\begin{algorithmic}[1]
\State \textbf{Input:} Dataset $\mathcal{D}$, base model $\theta^0$, token selection ratio $\rho$, balance coefficient $\gamma$.
\State Initialize history model $\theta_{\text{his}} \gets \theta^0$
\For{$t = 1$ \textbf{to} $T$}
    \State Sample a sequence $\boldsymbol{x}$ from $\mathcal{D}$
    \State Compute $\text{REL}(x_i)$ for each token via Eq.~\ref{eq:rel}
    \State Compute attention score $\text{Attn}(x_i)$ via Eq.~\ref{eq:attnscore}
    \State Normalize $\text{REL}(x_i)$ via Eq.~\ref{eq:norm}
    \State Calculate combined score:
    \begin{equation*}
       \text{Score}(x_i) = \gamma \cdot \text{Normalize}(\text{REL}(x_i)) + (1-\gamma) \cdot \text{AttnScore}(x_i)
     \end{equation*}
    \State Select top-$\rho$ tokens by $\text{Score}(x_i)$
    \State Update $\theta^t$ with selected tokens
    \State \textbf{Optional:} Update $\theta_{\text{his}}$ (e.g., EMA, Eq.~\ref{eq:ema})
\EndFor
\State \textbf{Return:} Fine-tuned model $\theta^t$
\end{algorithmic}
\label{alg}
\end{algorithm}

\section{Case Studies of Token Selection}
In this section, we present several case studies to illustrate the outcomes of different token selection strategies. 
For each sample, the tokens selected by a given method are highlighted in \textcolor{blue}{blue}. 
In particular, for \textsc{ssToken}, we additionally use \textcolor{orange}{orange} to mark tokens that are primarily selected due to their high attention scores, distinguishing them from those selected mainly by loss-based signals. 
These visualizations provide an intuitive understanding of how various approaches behave and highlight the complementary role of semantic-aware signals in our method.
\clearpage
\begin{tcolorbox}[colframe=gray!70!black,colback=white, title=Sample 1]
\textbf{User:}
Write a response that appropriately completes the request.

\#\#\# Instruction:

Name one type of organic compound.

\#\#\# Response:
\vspace{0.1in}\\
\textbf{Assistant (RHO-1):} \textcolor{blue}{One}\textcolor{blue}{~type}~of~organic\textcolor{blue}{~compound}\textcolor{blue}{~is}~an\textcolor{blue}{~alk}ane\textcolor{blue}{,}\textcolor{blue}{~which}~is\textcolor{blue}{~a}~saturated\textcolor{blue}{~hydro}carbon\textcolor{blue}{~that}\textcolor{blue}{~contains}~single~bonds\textcolor{blue}{~between}\textcolor{blue}{~carbon}\textcolor{blue}{~atoms}\textcolor{blue}{.}
\vspace{0.1in}\\
\textbf{Assistant (TokenCleaning):}
\textcolor{blue}{One}\textcolor{blue}{~type}~of~organic~compound~is~an\textcolor{blue}{~alk}ane\textcolor{blue}{,}\textcolor{blue}{~which}~is\textcolor{blue}{~a}~saturated\textcolor{blue}{~hydro}carbon\textcolor{blue}{~that}\textcolor{blue}{~contains}~single~bonds~between\textcolor{blue}{~carbon}~atoms\textcolor{blue}{.}
\vspace{0.1in}\\
\textbf{Assistant (ssToken):}
\textcolor{blue}{One}\textcolor{orange}{~type}~of\textcolor{orange}{~organic}\textcolor{orange}{~compound}\textcolor{orange}{~is}~an\textcolor{orange}{~alk}\textcolor{blue}{ane}\textcolor{orange}{,}\textcolor{orange}{~which}\textcolor{orange}{~is}~a~saturated~hydrocarbon\textcolor{orange}{~that}\textcolor{blue}{~contains}~single~bonds\textcolor{blue}{~between}\textcolor{blue}{~carbon}~atoms\textcolor{blue}{.}

\end{tcolorbox}

\begin{tcolorbox}[colframe=gray!70!black,colback=white, title=Sample 2]
\textbf{User:}
Provide instructions on how to accurately perform a bicep curl.
\vspace{0.1in}\\
\textbf{Assistant (RHO-1):}
 \textcolor{blue}{1} \textcolor{blue}{.} \textcolor{blue}{~Stand} \textcolor{blue}{~with} \textcolor{blue}{~your} \textcolor{blue}{~feet} \textcolor{blue}{~shoulder} \textcolor{blue}{-width} \textcolor{blue}{~apart} \textcolor{blue}{~and}~grab  \textcolor{blue}{~a} \textcolor{blue}{~dumb} \textcolor{blue}{bell}~in \textcolor{blue}{~each} \textcolor{blue}{~hand} \textcolor{blue}{~with} \textcolor{blue}{~your} \textcolor{blue}{~palms} \textcolor{blue}{~facing} \textcolor{blue}{~forward} \textcolor{blue}{.
} \textcolor{blue}{2}. \textcolor{blue}{~Keep} \textcolor{blue}{~your} \textcolor{blue}{~elbows} \textcolor{blue}{~close}~to ~your \textcolor{blue}{~body} \textcolor{blue}{~and}~slowly~lift~the~weights \textcolor{blue}{~towards} \textcolor{blue}{~your} \textcolor{blue}{~shoulders} \textcolor{blue}{,}~while~exhaling \textcolor{blue}{.}~Remember \textcolor{blue}{~to} \textcolor{blue}{~keep}  \textcolor{blue}{~your}~wrists \textcolor{blue}{~straight} \textcolor{blue}{~and} \textcolor{blue}{~engage} \textcolor{blue}{~your}  \textcolor{blue}{~b}iceps~to \textcolor{blue}{~lift} \textcolor{blue}{~the}~weight \textcolor{blue}{.
} \textcolor{blue}{3}.~Pause \textcolor{blue}{~for}~a~second \textcolor{blue}{~at}~the~top \textcolor{blue}{~of} \textcolor{blue}{~the} \textcolor{blue}{~curl} \textcolor{blue}{,} \textcolor{blue}{~squeezing} \textcolor{blue}{~your} \textcolor{blue}{~b} \textcolor{blue}{iceps} \textcolor{blue}{.
} \textcolor{blue}{4}. \textcolor{blue}{~Slow}ly \textcolor{blue}{~lower}~the \textcolor{blue}{~weights}~back~down~to~the~starting~position, \textcolor{blue}{~while}~inhaling \textcolor{blue}{.
} \textcolor{blue}{5}. \textcolor{blue}{~Repeat} \textcolor{blue}{~the} \textcolor{blue}{~exercise} \textcolor{blue}{~for}~the \textcolor{blue}{~desired} \textcolor{blue}{~number}~of \textcolor{blue}{~reps}.
 \textcolor{blue}{6}. \textcolor{blue}{~To} \textcolor{blue}{~avoid}~injury \textcolor{blue}{,} \textcolor{blue}{~avoid}~swinging \textcolor{blue}{~your}~arms \textcolor{blue}{~or}~using~momentum \textcolor{blue}{~to} \textcolor{blue}{~lift}~the~weights \textcolor{blue}{.}~Keep \textcolor{blue}{~your}~core~engaged \textcolor{blue}{~and}~focus~on~isolating~your~b \textcolor{blue}{iceps}~for \textcolor{blue}{~maximum} \textcolor{blue}{~benefit} \textcolor{blue}{.}
\vspace{0.1in}\\
\textbf{Assistant (TokenCleaning):}
 \textcolor{blue}{1} \textcolor{blue}{.} \textcolor{blue}{~Stand} \textcolor{blue}{~with} \textcolor{blue}{~your} \textcolor{blue}{~feet} \textcolor{blue}{~shoulder} \textcolor{blue}{-width}~apart \textcolor{blue}{~and}~grab \textcolor{blue}{~a} \textcolor{blue}{~dumb} \textcolor{blue}{bell}~in \textcolor{blue}{~each}~hand \textcolor{blue}{~with} \textcolor{blue}{~your} \textcolor{blue}{~palms}~facing \textcolor{blue}{~forward} \textcolor{blue}{.
} \textcolor{blue}{2}. \textcolor{blue}{~Keep}~your \textcolor{blue}{~elbows} \textcolor{blue}{~close}~to~your \textcolor{blue}{~body} \textcolor{blue}{~and}~slowly~lift~the~weights \textcolor{blue}{~towards} \textcolor{blue}{~your}~shoulders \textcolor{blue}{,}~while~exhaling \textcolor{blue}{.}~Remember \textcolor{blue}{~to} \textcolor{blue}{~keep} \textcolor{blue}{~your}~wrists~straight \textcolor{blue}{~and} \textcolor{blue}{~engage}~your \textcolor{blue}{~b}iceps~to \textcolor{blue}{~lift} \textcolor{blue}{~the}~weight \textcolor{blue}{.
} \textcolor{blue}{3}.~Pause \textcolor{blue}{~for}~a~second \textcolor{blue}{~at}~the~top \textcolor{blue}{~of} \textcolor{blue}{~the} \textcolor{blue}{~curl} \textcolor{blue}{,} \textcolor{blue}{~squeezing}~your \textcolor{blue}{~b} \textcolor{blue}{iceps} \textcolor{blue}{.
}4. \textcolor{blue}{~Slow}ly \textcolor{blue}{~lower}~the \textcolor{blue}{~weights}~back~down~to~the~starting~position, \textcolor{blue}{~while}~inhaling \textcolor{blue}{.
} \textcolor{blue}{5}. \textcolor{blue}{~Repeat} \textcolor{blue}{~the} \textcolor{blue}{~exercise} \textcolor{blue}{~for}~the \textcolor{blue}{~desired} \textcolor{blue}{~number}~of \textcolor{blue}{~reps}.
 \textcolor{blue}{6}. \textcolor{blue}{~To} \textcolor{blue}{~avoid}~injury \textcolor{blue}{,} \textcolor{blue}{~avoid}~swinging~your~arms~or~using~momentum \textcolor{blue}{~to} \textcolor{blue}{~lift}~the~weights \textcolor{blue}{.}~Keep \textcolor{blue}{~your}~core~engaged \textcolor{blue}{~and}~focus~on~isolating~your~b \textcolor{blue}{iceps}~for \textcolor{blue}{~maximum} \textcolor{blue}{~benefit} \textcolor{blue}{.}
\vspace{0.1in}\\
\textbf{Assistant (ssToken):}
 \textcolor{blue}{1} \textcolor{blue}{.}\textcolor{orange}{~Stand}\textcolor{orange}{~with} \textcolor{blue}{~your}\textcolor{orange}{~feet}\textcolor{orange}{~shoulder}-width \textcolor{blue}{~apart} \textcolor{blue}{~and}\textcolor{orange}{~grab}\textcolor{orange}{~a}\textcolor{orange}{~dumb} \textcolor{blue}{bell}\textcolor{orange}{~in} \textcolor{blue}{~each} \textcolor{blue}{~hand} \textcolor{blue}{~with} \textcolor{blue}{~your}\textcolor{orange}{~palms}\textcolor{orange}{~facing}~forward.
 \textcolor{blue}{2} \textcolor{blue}{.}\textcolor{orange}{~Keep} \textcolor{blue}{~your}\textcolor{orange}{~elbows}~close \textcolor{blue}{~to}~your~body~and\textcolor{orange}{~slowly} \textcolor{blue}{~lift}~the \textcolor{blue}{~weights} \textcolor{blue}{~towards} \textcolor{blue}{~your} \textcolor{blue}{~shoulders} \textcolor{blue}{,} \textcolor{blue}{~while}\textcolor{orange}{~exh} \textcolor{blue}{aling}.~Remember \textcolor{blue}{~to}\textcolor{orange}{~keep} \textcolor{blue}{~your} \textcolor{blue}{~wrists}~straight \textcolor{blue}{~and} \textcolor{blue}{~engage} \textcolor{blue}{~your} \textcolor{blue}{~b} \textcolor{blue}{iceps} \textcolor{blue}{~to} \textcolor{blue}{~lift}~the~weight.
 \textcolor{blue}{3}.\textcolor{orange}{~Pause} \textcolor{blue}{~for}~a \textcolor{blue}{~second} \textcolor{blue}{~at}~the~top \textcolor{blue}{~of}~the \textcolor{blue}{~curl} \textcolor{blue}{,} \textcolor{blue}{~squeezing}~your~biceps.
 \textcolor{blue}{4}.\textcolor{orange}{~Slow} \textcolor{blue}{ly} \textcolor{blue}{~lower}~the~weights~back~down \textcolor{blue}{~to}~the~starting~position,~while\textcolor{orange}{~inh}aling.
 \textcolor{blue}{5}. \textcolor{blue}{~Repeat}~the \textcolor{blue}{~exercise} \textcolor{blue}{~for}~the \textcolor{blue}{~desired}~number~of \textcolor{blue}{~reps}.
 \textcolor{blue}{6} \textcolor{blue}{.}~To \textcolor{blue}{~avoid}\textcolor{orange}{~injury} \textcolor{blue}{,} \textcolor{blue}{~avoid}\textcolor{orange}{~swinging}~your \textcolor{blue}{~arms} \textcolor{blue}{~or}\textcolor{orange}{~using}\textcolor{orange}{~momentum} \textcolor{blue}{~to} \textcolor{blue}{~lift}~the~weights \textcolor{blue}{.} \textcolor{blue}{~Keep} \textcolor{blue}{~your} \textcolor{blue}{~core}~engaged~and~focus \textcolor{blue}{~on} \textcolor{blue}{~isol} \textcolor{blue}{ating}~your \textcolor{blue}{~b}iceps~for \textcolor{blue}{~maximum}~benefit \textcolor{blue}{.}
\end{tcolorbox}

\begin{tcolorbox}[colframe=gray!70!black,colback=white, title=Sample 3]
\textbf{User:}
In a restaurant kitchen, a chef uses a food processor to prepare a sauce. The recipe calls for 2 cups of chopped tomatoes, 1 cup of diced onions, and 1/2 cup of minced garlic. The food processor has a maximum capacity of 4 cups. If the chef wants to make enough sauce for 6 servings, how many times must the ingredients be processed in the food processor? 
Hint: Assume that the food processor can process the entire recipe at once. \vspace{0.1in}
\\
\textbf{Assistant (RHO-1):}
\textcolor{blue}{The}\textcolor{blue}{~recipe}\textcolor{blue}{~requires}\textcolor{blue}{~a}\textcolor{blue}{~total}\textcolor{blue}{~of}\textcolor{blue}{~}2~+\textcolor{blue}{~}\textcolor{blue}{1}\textcolor{blue}{~+}\textcolor{blue}{~}\textcolor{blue}{0}\textcolor{blue}{.}\textcolor{blue}{5}~=\textcolor{blue}{~}\textcolor{blue}{3}\textcolor{blue}{.}5\textcolor{blue}{~cups}\textcolor{blue}{~of}\textcolor{blue}{~ingredients}\textcolor{blue}{.}\textcolor{blue}{~To}\textcolor{blue}{~make}\textcolor{blue}{~enough}\textcolor{blue}{~sauce}\textcolor{blue}{~for}\textcolor{blue}{~}\textcolor{blue}{6}\textcolor{blue}{~servings},\textcolor{blue}{~the}~chef\textcolor{blue}{~needs}~to~prepare\textcolor{blue}{~}\textcolor{blue}{6}~*\textcolor{blue}{~}3.5~=\textcolor{blue}{~}\textcolor{blue}{21}\textcolor{blue}{~cups}\textcolor{blue}{~of}~ingredients~in~total\textcolor{blue}{.}\textcolor{blue}{~
}\textcolor{blue}{Since}\textcolor{blue}{~the}\textcolor{blue}{~food}~processor~can\textcolor{blue}{~process}~up~to\textcolor{blue}{~}4~cups~of\textcolor{blue}{~ingredients}\textcolor{blue}{~at}~a~time\textcolor{blue}{,}\textcolor{blue}{~the}\textcolor{blue}{~chef}~needs\textcolor{blue}{~to}\textcolor{blue}{~process}\textcolor{blue}{~the}\textcolor{blue}{~ingredients}~21~/~\textcolor{blue}{4}\textcolor{blue}{~=}\textcolor{blue}{~}\textcolor{blue}{5}.25\textcolor{blue}{~times}\textcolor{blue}{.}~
\textcolor{blue}{However},~since\textcolor{blue}{~the}~chef~cannot\textcolor{blue}{~process}\textcolor{blue}{~ingredients}~a~fraction~of\textcolor{blue}{~a}~time,~they\textcolor{blue}{~will}\textcolor{blue}{~need}\textcolor{blue}{~to}\textcolor{blue}{~process}\textcolor{blue}{~the}~ingredients~6\textcolor{blue}{~times}~to~make\textcolor{blue}{~enough}~sauce~for\textcolor{blue}{~}\textcolor{blue}{6}\textcolor{blue}{~servings}\textcolor{blue}{.}
\vspace{0.1in}\\
\textbf{Assistant (TokenCleaning):}
\textcolor{blue}{The}\textcolor{blue}{~recipe}\textcolor{blue}{~requires}\textcolor{blue}{~a}\textcolor{blue}{~total}\textcolor{blue}{~of}\textcolor{blue}{~}2~+\textcolor{blue}{~}\textcolor{blue}{1}\textcolor{blue}{~+}\textcolor{blue}{~}\textcolor{blue}{0}\textcolor{blue}{.}\textcolor{blue}{5}~=\textcolor{blue}{~}\textcolor{blue}{3}\textcolor{blue}{.}\textcolor{blue}{5}\textcolor{blue}{~cups}\textcolor{blue}{~of}\textcolor{blue}{~ingredients}\textcolor{blue}{.}\textcolor{blue}{~To}\textcolor{blue}{~make}\textcolor{blue}{~enough}\textcolor{blue}{~sauce}\textcolor{blue}{~for}\textcolor{blue}{~}\textcolor{blue}{6}\textcolor{blue}{~servings}\textcolor{blue}{,}\textcolor{blue}{~the}~chef\textcolor{blue}{~needs}~to~prepare\textcolor{blue}{~}\textcolor{blue}{6}~*\textcolor{blue}{~}3\textcolor{blue}{.}\textcolor{blue}{5}~=\textcolor{blue}{~}\textcolor{blue}{21}\textcolor{blue}{~cups}\textcolor{blue}{~of}~ingredients~in~total\textcolor{blue}{.}\textcolor{blue}{~
}\textcolor{blue}{Since}\textcolor{blue}{~the}\textcolor{blue}{~food}\textcolor{blue}{~processor}~can\textcolor{blue}{~process}\textcolor{blue}{~up}\textcolor{blue}{~to}\textcolor{blue}{~}\textcolor{blue}{4}\textcolor{blue}{~cups}~of\textcolor{blue}{~ingredients}\textcolor{blue}{~at}~a\textcolor{blue}{~time}\textcolor{blue}{,}\textcolor{blue}{~the}\textcolor{blue}{~chef}~needs\textcolor{blue}{~to}\textcolor{blue}{~process}\textcolor{blue}{~the}\textcolor{blue}{~ingredients}~21~/\textcolor{blue}{~}\textcolor{blue}{4}\textcolor{blue}{~=}\textcolor{blue}{~}\textcolor{blue}{5}.25\textcolor{blue}{~times}\textcolor{blue}{.}~
\textcolor{blue}{However}\textcolor{blue}{,}~since\textcolor{blue}{~the}~chef~cannot\textcolor{blue}{~process}\textcolor{blue}{~ingredients}~a~fraction~of\textcolor{blue}{~a}~time,~they\textcolor{blue}{~will}\textcolor{blue}{~need}\textcolor{blue}{~to}\textcolor{blue}{~process}\textcolor{blue}{~the}~ingredients~6\textcolor{blue}{~times}~to~make\textcolor{blue}{~enough}\textcolor{blue}{~sauce}~for\textcolor{blue}{~}\textcolor{blue}{6}\textcolor{blue}{~servings}\textcolor{blue}{.}
\vspace{0.1in}\\
\textbf{Assistant (ssToken):}
 \textcolor{orange}{The}\textcolor{blue}{~recipe}\textcolor{orange}{~requires}\textcolor{blue}{~a}\textcolor{orange}{~total}\textcolor{orange}{~of}\textcolor{blue}{~}2~\textcolor{blue}{+}~\textcolor{blue}{1}\textcolor{orange}{~+}~\textcolor{blue}{0.}\textcolor{blue}{5}~=~\textcolor{blue}{3}.\textcolor{blue}{5}\textcolor{orange}{~cups}\textcolor{orange}{~of}\textcolor{orange}{~ingredients}\textcolor{orange}{.}\textcolor{orange}{~To}\textcolor{blue}{~make}\textcolor{blue}{~enough}\textcolor{orange}{~sauce}\textcolor{blue}{~for}~\textcolor{blue}{6}\textcolor{orange}{~servings}\textcolor{orange}{,}~the\textcolor{orange}{~chef}\textcolor{blue}{~needs}\textcolor{blue}{~to}\textcolor{blue}{~prepare}~\textcolor{blue}{6}~*~3\textcolor{blue}{.}5\textcolor{blue}{~=}\textcolor{blue}{~}21\textcolor{blue}{~cups}\textcolor{blue}{~of}~ingredients~in\textcolor{orange}{~total}\textcolor{orange}{.}\textcolor{orange}{~Since}\textcolor{blue}{~the}\textcolor{blue}{~food}\textcolor{orange}{~processor}~can~process~up~to~4~cups~of\textcolor{blue}{~ingredients}\textcolor{blue}{~at}~a\textcolor{orange}{~time}\textcolor{orange}{,}~the\textcolor{orange}{~chef}\textcolor{orange}{~needs}\textcolor{orange}{~to}\textcolor{orange}{~process}~the\textcolor{blue}{~ingredients}~\textcolor{orange}{21}\textcolor{blue}{~/}~4\textcolor{blue}{~=}~5.\textcolor{blue}{25}\textcolor{blue}{~times}\textcolor{blue}{.}~However\textcolor{orange}{,}\textcolor{orange}{~since}~the\textcolor{orange}{~chef}\textcolor{blue}{~cannot}\textcolor{orange}{~process}\textcolor{blue}{~ingredients}~a\textcolor{blue}{~fraction}\textcolor{orange}{~of}~a~time\textcolor{orange}{,}\textcolor{blue}{~they}\textcolor{blue}{~will}\textcolor{blue}{~need}\textcolor{blue}{~to}\textcolor{blue}{~process}~the\textcolor{blue}{~ingredients}~6\textcolor{blue}{~times}\textcolor{orange}{~to}~make~enough\textcolor{blue}{~sauce}~for~\textcolor{blue}{6}\textcolor{blue}{~servings}\textcolor{blue}{.}
\end{tcolorbox}

\begin{tcolorbox}[colframe=gray!70!black,colback=white, title=Sample 4]
\textbf{User:}
Using a deterministic algorithm, please generate a set of 3-6 integers, all of which must fall within the inclusive range of 10 to 100. These integers must satisfy the following conditions: the sum of all numbers in the set must be greater than or equal to 200, the difference between the largest and smallest numbers must be greater than or equal to 50, and no two numbers in the set may have a common factor greater than 1. Please provide the resulting set of integers.
\vspace{0.1in}\\
\textbf{Assistant (RHO-1):} \textcolor{blue}{Here}\textcolor{blue}{~is}\textcolor{blue}{~a}\textcolor{blue}{~set}\textcolor{blue}{~of}\textcolor{blue}{~integers}\textcolor{blue}{~that}~satisfies\textcolor{blue}{~the}\textcolor{blue}{~given}\textcolor{blue}{~conditions}\textcolor{blue}{:}~-~\textcolor{blue}{13}\textcolor{blue}{,}~31\textcolor{blue}{,}\textcolor{blue}{~}97\textcolor{blue}{,}\textcolor{blue}{~}49~The\textcolor{blue}{~sum}\textcolor{blue}{~of}~these\textcolor{blue}{~numbers}\textcolor{blue}{~is}\textcolor{blue}{~}\textcolor{blue}{190}\textcolor{blue}{,}\textcolor{blue}{~which}\textcolor{blue}{~is}~less\textcolor{blue}{~than}~\textcolor{blue}{200}\textcolor{blue}{,}~so\textcolor{blue}{~we}~can\textcolor{blue}{~add}\textcolor{blue}{~another}\textcolor{blue}{~number}\textcolor{blue}{~to}\textcolor{blue}{~the}\textcolor{blue}{~set}\textcolor{blue}{.}\textcolor{blue}{~-}\textcolor{blue}{~}\textcolor{blue}{11}\textcolor{blue}{,}~\textcolor{blue}{29},\textcolor{blue}{~}\textcolor{blue}{89},~47,\textcolor{blue}{~}\textcolor{blue}{35}\textcolor{blue}{~This}~set~satisfies\textcolor{blue}{~all}\textcolor{blue}{~the}~given~conditions\textcolor{blue}{.}\textcolor{blue}{~The}\textcolor{blue}{~sum}~of~the~numbers\textcolor{blue}{~is}~\textcolor{blue}{211}\textcolor{blue}{,}~which~is\textcolor{blue}{~greater}~than\textcolor{blue}{~or}~equal~to\textcolor{blue}{~}200.\textcolor{blue}{~The}\textcolor{blue}{~difference}~between~the\textcolor{blue}{~largest}~and~smallest~numbers\textcolor{blue}{~is}~\textcolor{blue}{78}\textcolor{blue}{,}\textcolor{blue}{~which}~is\textcolor{blue}{~greater}~than~or\textcolor{blue}{~equal}\textcolor{blue}{~to}~50\textcolor{blue}{.}~Finally,~no\textcolor{blue}{~two}\textcolor{blue}{~numbers}~in\textcolor{blue}{~the}\textcolor{blue}{~set}\textcolor{blue}{~have}\textcolor{blue}{~a}\textcolor{blue}{~common}\textcolor{blue}{~factor}~greater~than\textcolor{blue}{~}1\textcolor{blue}{.}
\vspace{0.1in}\\
\textbf{Assistant (TokenCleaning):}
\textcolor{blue}{Here}\textcolor{blue}{~is}\textcolor{blue}{~a}\textcolor{blue}{~set}\textcolor{blue}{~of}\textcolor{blue}{~integers}\textcolor{blue}{~that}~satisfies\textcolor{blue}{~the}\textcolor{blue}{~given}\textcolor{blue}{~conditions}\textcolor{blue}{:}~-~\textcolor{blue}{13}\textcolor{blue}{,}~31\textcolor{blue}{,}\textcolor{blue}{~}97\textcolor{blue}{,}\textcolor{blue}{~}49~The\textcolor{blue}{~sum}\textcolor{blue}{~of}~these\textcolor{blue}{~numbers}\textcolor{blue}{~is}\textcolor{blue}{~}\textcolor{blue}{190}\textcolor{blue}{,}\textcolor{blue}{~which}\textcolor{blue}{~is}~less\textcolor{blue}{~than}~\textcolor{blue}{200}\textcolor{blue}{,}~so\textcolor{blue}{~we}~can\textcolor{blue}{~add}\textcolor{blue}{~another}\textcolor{blue}{~number}\textcolor{blue}{~to}\textcolor{blue}{~the}\textcolor{blue}{~set}\textcolor{blue}{.}\textcolor{blue}{~-}\textcolor{blue}{~}\textcolor{blue}{11}\textcolor{blue}{,}~\textcolor{blue}{29},~\textcolor{blue}{89},~47,~\textcolor{blue}{35}\textcolor{blue}{~This}~set~satisfies\textcolor{blue}{~all}\textcolor{blue}{~the}~given~conditions\textcolor{blue}{.}\textcolor{blue}{~The}\textcolor{blue}{~sum}~of~the~numbers\textcolor{blue}{~is}~\textcolor{blue}{211}\textcolor{blue}{,}~which~is\textcolor{blue}{~greater}~than\textcolor{blue}{~or}~equal~to\textcolor{blue}{~}200.\textcolor{blue}{~The}\textcolor{blue}{~difference}~between~the\textcolor{blue}{~largest}~and~smallest~numbers~is~\textcolor{blue}{78}\textcolor{blue}{,}\textcolor{blue}{~which}~is\textcolor{blue}{~greater}~than~or\textcolor{blue}{~equal}~to~50\textcolor{blue}{.}~Finally,~no\textcolor{blue}{~two}~numbers~in\textcolor{blue}{~the}\textcolor{blue}{~set}\textcolor{blue}{~have}\textcolor{blue}{~a}\textcolor{blue}{~common}~factor~greater~than\textcolor{blue}{~}1\textcolor{blue}{.}
\vspace{0.1in}\\
\textbf{Assistant (ssToken):}
\textcolor{blue}{Here}\textcolor{orange}{~is}\textcolor{blue}{~a}~set\textcolor{orange}{~of}\textcolor{blue}{~integers}\textcolor{orange}{~that}\textcolor{orange}{~satisfies}\textcolor{blue}{~the}\textcolor{blue}{~given}\textcolor{orange}{~conditions}\textcolor{orange}{:}~-~\textcolor{orange}{13}\textcolor{blue}{,}\textcolor{blue}{~}\textcolor{orange}{31},~97\textcolor{blue}{,}~\textcolor{orange}{49}~The\textcolor{orange}{~sum}\textcolor{orange}{~of}~these\textcolor{orange}{~numbers}\textcolor{orange}{~is}\textcolor{blue}{~}\textcolor{blue}{190}\textcolor{orange}{,}\textcolor{blue}{~which}\textcolor{blue}{~is}~less\textcolor{orange}{~than}\textcolor{blue}{~}\textcolor{orange}{200}\textcolor{blue}{,}\textcolor{orange}{~so}\textcolor{blue}{~we}~can\textcolor{blue}{~add}\textcolor{blue}{~another}\textcolor{blue}{~number}\textcolor{orange}{~to}~the~set\textcolor{blue}{.}~-~11,\textcolor{orange}{~}29,~89,~47,\textcolor{orange}{~}35\textcolor{blue}{~This}\textcolor{blue}{~set}\textcolor{blue}{~satisfies}\textcolor{blue}{~all}~the~given\textcolor{orange}{~conditions}\textcolor{orange}{.}\textcolor{blue}{~The}\textcolor{blue}{~sum}\textcolor{blue}{~of}~the\textcolor{blue}{~numbers}\textcolor{blue}{~is}\textcolor{orange}{~}\textcolor{blue}{211}\textcolor{orange}{,}\textcolor{blue}{~which}~is~greater~than\textcolor{blue}{~or}~equal~to~\textcolor{blue}{200}\textcolor{blue}{.}\textcolor{orange}{~The}\textcolor{blue}{~difference}\textcolor{orange}{~between}\textcolor{blue}{~the}\textcolor{blue}{~largest}~and~smallest~numbers\textcolor{blue}{~is}\textcolor{blue}{~}78\textcolor{blue}{,}~which~is~greater~than~or~equal~to~\textcolor{blue}{50}\textcolor{blue}{.}\textcolor{orange}{~Finally}\textcolor{blue}{,}\textcolor{orange}{~no}\textcolor{blue}{~two}\textcolor{blue}{~numbers}\textcolor{orange}{~in}~the\textcolor{blue}{~set}\textcolor{orange}{~have}\textcolor{blue}{~a}\textcolor{blue}{~common}~factor~greater~than~\textcolor{orange}{1}\textcolor{blue}{.}
\end{tcolorbox}
\end{document}